# A Probabilistic Framework for Learning Kinematic Models of Articulated Objects

**Jürgen Sturm** STURM@INFORMATIK.UNI-FREIBURG.DE
**Cyrill Stachniss** STACHNIS@INFORMATIK.UNI-FREIBURG.DE
**Wolfram Burgard** BURGARD@INFORMATIK.UNI-FREIBURG.DE
*Department of Computer Science,*
*University of Freiburg,*
*Georges-Koehler-Allee 79, 79100 Freiburg, Germany*

## Abstract

Robots operating in domestic environments generally need to interact with articulated objects, such as doors, cabinets, dishwashers or fridges. In this work, we present a novel, probabilistic framework for modeling articulated objects as kinematic graphs. Vertices in this graph correspond to object parts, while edges between them model their kinematic relationship. In particular, we present a set of parametric and non-parametric edge models and how they can robustly be estimated from noisy pose observations. We furthermore describe how to estimate the kinematic structure and how to use the learned kinematic models for pose prediction and for robotic manipulation tasks. We finally present how the learned models can be generalized to new and previously unseen objects. In various experiments using real robots with different camera systems as well as in simulation, we show that our approach is valid, accurate and efficient. Further, we demonstrate that our approach has a broad set of applications, in particular for the emerging fields of mobile manipulation and service robotics.

## 1. Introduction

Service robots operating in domestic environments are typically faced with a variety of objects they have to deal with or they have to manipulate to fulfill their task. A further complicating factor is that many of the relevant objects are articulated, such as doors, windows, but also pieces of furniture like cupboards, cabinets, or larger objects such as garage doors, gates and cars. Understanding the spatial movements of the individual parts of articulated objects is essential for service robots to allow them to plan relevant actions such as door-opening trajectories and to assess whether they actually were successful. In this work, we investigate the problem of learning kinematic models of articulated objects and using them for robotic manipulation tasks. As an illustrating example, consider Fig. 1 where a mobile manipulation robot interacts with various articulated objects in a kitchen environment, learns their kinematic properties and infers their kinematic structure.

Our problem can be formulated as follows: Given a sequence of pose observations of object parts, our goal is to learn a compact kinematic model describing the *whole* articulated object. This kinematic model has to define (i) the connectivity between the parts, (ii) the number of degrees of freedom of the object, and (iii) the kinematic function for the articulated object. As a result, we obtain a generative model which can be used by the robot for generating and reasoning about future or unseen configurations.

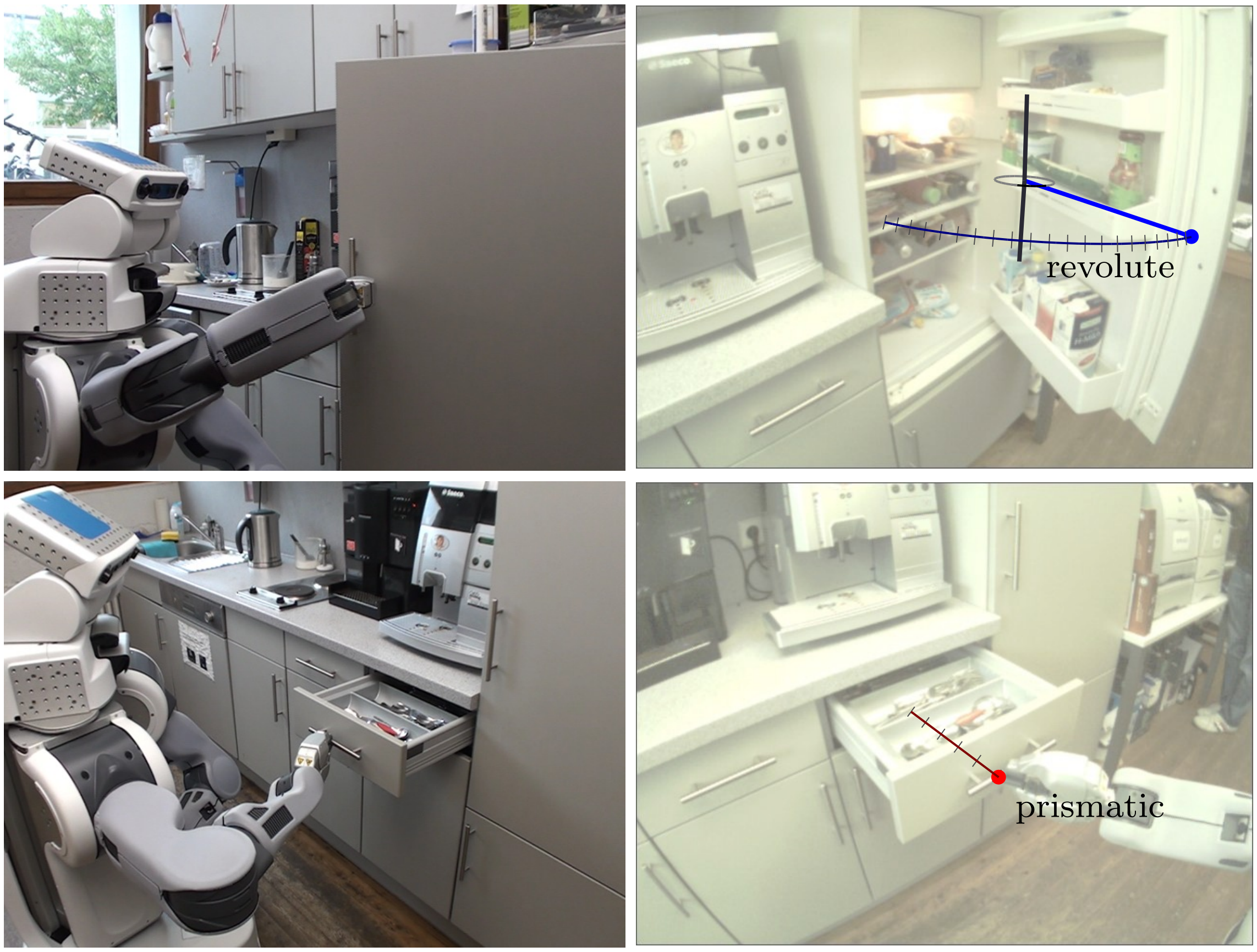

Figure 1: A service robot learns kinematic models of articulated objects in a kitchen environment.

The contribution of this paper is a novel approach that enables a real robot to learn kinematic models of articulated objects from its sensor data. These models describe the kinematics of the object and include the part connectivity, degrees of freedom of the objects, and kinematic constraints. We utilize these models subsequently to control the motion of the manipulator. Furthermore, we show how a robot can improve model learning by exploiting past experience. Finally, we show how our framework can be generalized to deal with closed-chain objects, i.e., objects that contain kinematic loops.

In the past, several researchers have addressed the problem to handle doors and drawers (Jain & Kemp, 2009a; Klingbeil, Saxena, & Ng, 2009; Meeussen et al., 2010; Wieland, Gonzalez-Aguirre, Vahrenkamp, Asfour, & Dillmann, 2009; McClung, Zheng, & Morrell, 2010). Most of these approaches, however, are either entirely model-free or assume substantial knowledge about the model and its parameters. Whereas model-free approaches release designers from providing any a-priori model information, the knowledge about objects and their articulation properties supports state estimation, motion prediction, and planning.

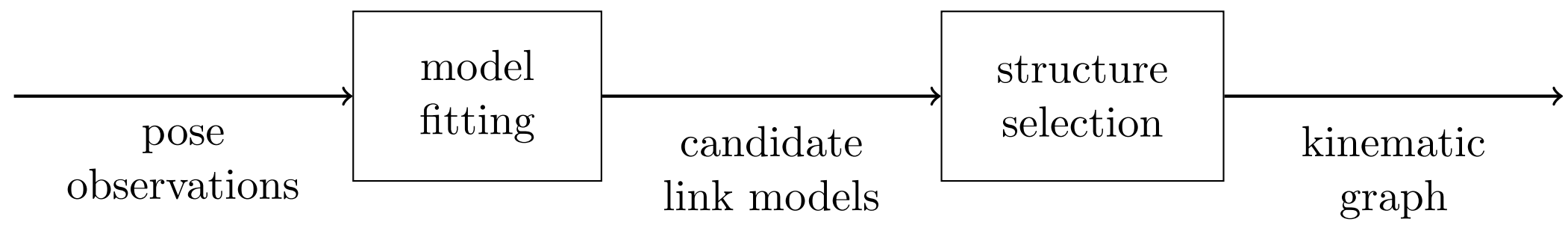

Figure 2: Schematic overview of our approach. The robot observes an articulated object in different poses. It uses these observations to generate a set of candidate models, and selects the kinematic structure that maximizes the posterior probability.

In our previous work, we introduced a simpler version of our probabilistic framework for modeling articulated objects, presented estimators for fitting link models, and showed how to efficiently find the kinematic structure for kinematic trees (Sturm, Pradeep, Stachniss, Plagemann, Konolige, & Burgard, 2009). As observations, we used a motion capture studio and data from simulation. Further, we used a stereo camera system for learning models for kitchen furniture (Sturm, Konolige, Stachniss, & Burgard, 2010). We described how a manipulation robot can learn kinematic models from direct interaction with articulated objects, and can improve over time by learning from experience (Sturm, Jain, Stachniss, Kemp, & Burgard, 2010). In this work, we present a unified framework for learning kinematic models of articulated objects with an extended set of experiments. In contrast to our previous work, we generalize our approach from kinematic trees to general kinematic graph objects and add a strategy to efficiently find a locally optimal graph. With this, it becomes possible to model articulated objects that contain kinematic loops. Furthermore, finding the effective number of degrees of freedom (DOFs) of an articulated object directly follows from our approach. A general software framework that implements the presented approach is available online[1] under the BSD license, including source code, documentation, and tutorials.

This paper is organized as follows. In Section 2, we introduce our unified framework for modeling the kinematics of articulated objects. In Section 3, we present several extensions including the exploitation of prior information, kinematic loops, and the estimation of degrees of freedom. In Section 4, we describe different options to perceive and control the motion of articulated objects. We analyze our approach in an extensive set of experiments both in simulation and on real robots and report our results in Section 5. Finally, we conclude this article with a discussion of related work in Section 6.

## 2. A Probabilistic Framework for Articulated Objects

We define an articulated object to consist of multiple object parts that have one or more passively actuated mechanical links between them. These links constrain the motion between the parts. For example, the hinge of a door constrains the door to move on an arc, and the shaft of a drawer constrains the drawer to move on a line segment. The simplest articulated object consists of two rigid parts with one mechanical link. More complex ob-

1. http://www.ros.org/wiki/articulation

jects may consist of several articulated parts, like a door with a door handle, or a car with several doors, windows, and wheels.

Fig. 2 gives a high-level overview of our proposed system. A robot observes the pose of an articulated object being manipulated. For the relative motion of any two parts, it fits different candidate models that describe different mechanical links. From this set of candidate link models, it selects the kinematic structure that best explains the observed motion, i.e., the kinematic structure that maximizes the posterior probability.

### 2.1 Notation

We assume that a robot, external to the object, observes the pose of an articulated object consisting of $p$ object parts. We denote the true pose of object part $i \in \{1, \dots, p\}$ by a vector $\mathbf{x}_i \in SE(3)$ representing the 3D pose of that part (including position and orientation), where $SE(3) = \mathbb{R}^3 \times SO(3)$ stands for the special Euclidean group. Further, we refer to the full object pose (containing the poses of all parts) with the vector $\mathbf{x}_{1:p} = (\mathbf{x}_1, \dots, \mathbf{x}_p)^T$. Two object parts $i$ and $j$ are related by their relative transformation $\boldsymbol{\Delta}_{ij} = \mathbf{x}_i \ominus \mathbf{x}_j$. We use $\oplus$ and $\ominus$ for referring to the motion composition operator and its inverse[2].

We denote a kinematic link model between two object parts $i$ and $j$ as $\mathcal{M}_{ij}$ and its associated parameter vector as $\boldsymbol{\theta}_{ij} \in \mathbb{R}^{k_{ij}}$, where $k_{ij} \in \mathbb{N}_0$ denotes the number of parameters of the model describing the link. A kinematic graph $G = (V_G, E_G)$ consists of a set of vertices $V_G = \{1, \dots, p\}$, corresponding to the parts of the articulated object, and a set of undirected edges $E_G \subset V_G \times V_G$ describing the kinematic link between two object parts. With each edge $(ij)$, a corresponding kinematic link model $\mathcal{M}_{ij}$ with parameter vector $\boldsymbol{\theta}_{ij}$ is associated.

All kinematic link models that we consider here (except for the trivial rigid link) have a latent variable $\mathbf{q}_{ij} \in C_{ij} \subset \mathbb{R}^{d_{ij}}$ that describes the configuration of the link. For a door, this can be the opening angle. $C_{ij}$ stands for the configuration space of the link. The variable $d_{ij}$ represents the number of DOFs of the mechanical link between the two parts.

While the object is being articulated, the robot observes the object pose; we denote the $n$-th pose observation of object part $i$ as $\mathbf{y}_i^n$. Correspondingly, we denote the $n$-th pose observation of all parts as $\mathbf{y}_{1:p}^n$ and a sequence of $n$ pose observations as $\mathcal{D}_{\mathbf{y}} = (\mathbf{y}_{1:p}^1, \dots, \mathbf{y}_{1:p}^n)$. Further, we will refer to $\mathcal{D}_{\mathbf{z}_{ij}} = (\mathbf{z}_{ij}^1, \dots, \mathbf{z}_{ij}^n)$ as the sequence of relative transformations $\mathbf{z}_{ij} = \mathbf{y}_i \ominus \mathbf{y}_j$ that the robot has observed so far for the edge $(ij)$.

Fig. 3a depicts a graphical model of a simple articulated object that consists of two object parts. We use here the so-called plate notation to simplify the notation of the graphical model. Here, the nodes inside the rectangle (the plate) are copied for $n$ times, i.e., for each time step $t$ in which the object is observed. In each of these time steps, the articulated object takes a particular configuration $\mathbf{q}_{12}$ defining – together with the model and its parameters – the noise-free relative transformation $\boldsymbol{\Delta}_{12}$ between the noise-free pose of the object parts $\mathbf{x}_1$ and $\mathbf{x}_2$. From that, the robot observes the noisy poses $\mathbf{y}_1$ and $\mathbf{y}_2$, and infers from them a virtual measurement $\mathbf{z}_{12} = \mathbf{y}_1 \ominus \mathbf{y}_2$. During model learning, the robot infers from these observations the link model $\mathcal{M}_{12}$ and link parameters $\boldsymbol{\theta}_{12}$.

2. E.g., if the poses $\mathbf{x}_1, \mathbf{x}_2 \in \mathbb{R}^{4\times 4}$ are represented as homogeneous matrices, then these operators correspond to matrix multiplication $\mathbf{x}_1 \oplus \mathbf{x}_2 = \mathbf{x}_1\mathbf{x}_2$ and inverse multiplication $\mathbf{x}_1 \ominus \mathbf{x}_2 = (\mathbf{x}_1)^{-1}\mathbf{x}_2$, respectively.

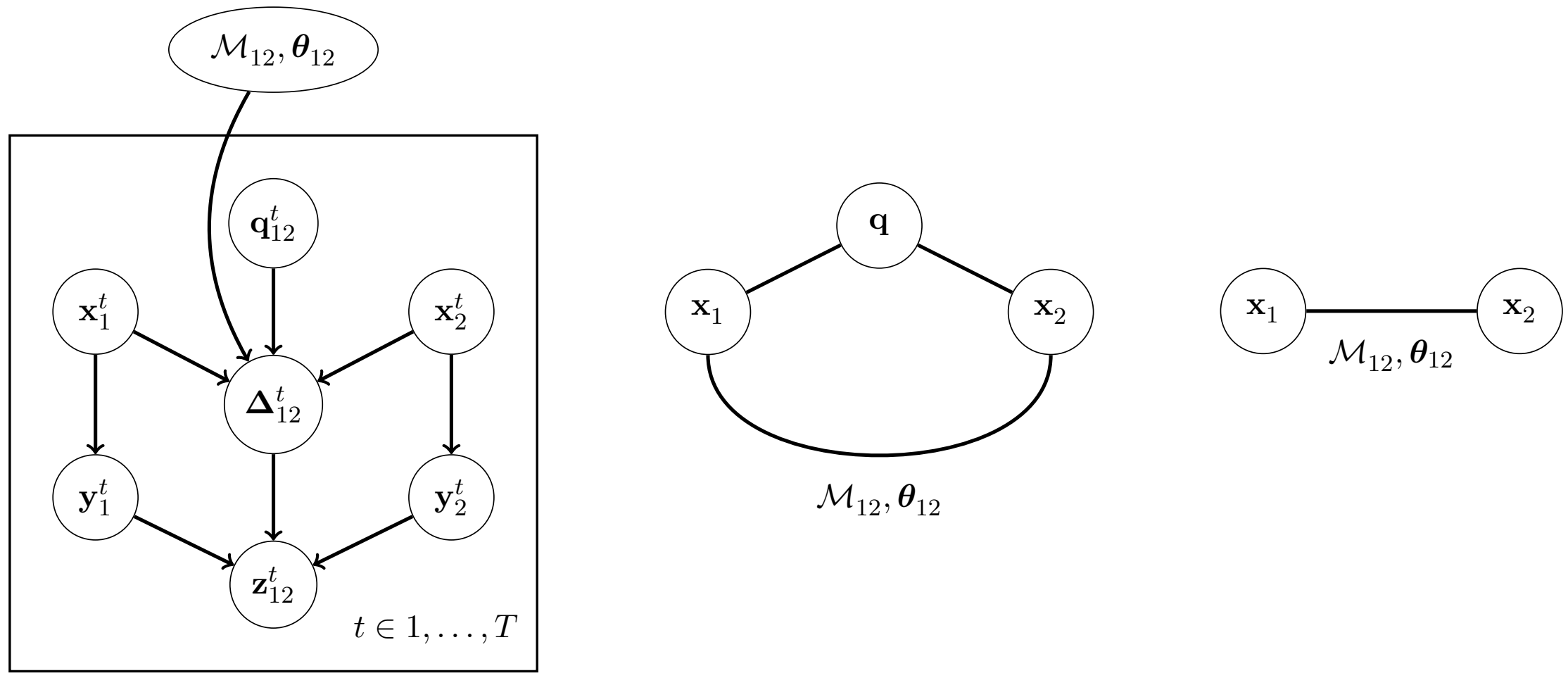

(a) Full graphical model (b) Reduced graphical model (c) Kinematic graph

Figure 3: (a) Graphical model of an articulated object consisting of two parts $\mathbf{x}_1$ and $\mathbf{x}_2$, being observed over $t$ time steps. The model $\mathcal{M}_{12}, \boldsymbol{\theta}_{12}$ is shared between all time steps. (b) shows a simplified version of the same graphical model. (c) shows the corresponding kinematic graph.

A reduced version of this graphical model is depicted in Fig. 3b. To improve readability, we leave out some nodes, i.e., the node corresponding to the relative transformation $\boldsymbol{\Delta}_{12}$ and the observation nodes $\mathbf{y}_1$, $\mathbf{y}_2$, and $\mathbf{z}_{12}$. Instead, we visualize the dependency between $\mathbf{x}_1$ and $\mathbf{x}_2$ by a direct link and label it with the corresponding model. Further, we collapse the configuration of the link into a single node corresponding to the configuration of the whole object. Finally, we refer to the kinematic graph as the graph that models the connectivity between object parts, as depicted in Fig. 3c).

### 2.2 Problem Definition

The problem that we consider here is to find the most likely kinematic graph $\hat{G}$ given a sequence of pose observations $\mathcal{D}_{\mathbf{y}}$ of an articulated object. In Bayesian terms, this means that we aim at finding the kinematic graph $\hat{G}$ that maximizes the posterior probability of observing the poses $\mathcal{D}_{\mathbf{y}}$ of the articulated object, i.e.,

$$\hat{G} = \arg\max_{G} p(G \mid \mathcal{D}_{\mathbf{y}}). \tag{1}$$

However, finding the global maximum of the posterior $p(G \mid \mathcal{D}_{\mathbf{y}})$ is difficult, because it is a highly non-convex function over a high-dimensional parameter space consisting of discrete as well as continuous dimensions that encode the kinematic structure and the kinematic properties, respectively.

Therefore, in this section, we consider a simplified problem. We restrict the structure space to kinematic trees only, and will focus on the general problem in Section 3. Kinematic trees have the property that their individual edges are independent of each other. As a result, we can estimate the link parameters independently of each other and also independent of the kinematic structure. This means that for learning the local kinematic relationship between object parts $i$ and $j$, only their relative transformations $\mathcal{D}_{\mathbf{z}_{ij}} = (\mathbf{z}_{ij}^1, \ldots, \mathbf{z}_{ij}^n)$ are relevant for estimating the edge model. With this, we can rephrase the maximization problem of (1) for kinematic trees now as

$$\hat{G} = \arg\max_G p(G \mid \mathcal{D}_{\mathbf{z}}) \tag{2}$$

$$= \arg\max_G p(\{(\mathcal{M}_{ij}, \boldsymbol{\theta}_{ij}) \mid (ij) \in E_G\} \mid \mathcal{D}_{\mathbf{z}}). \tag{3}$$

$$= \arg\max_G \prod_{(ij) \in E_G} p(\mathcal{M}_{ij}, \boldsymbol{\theta}_{ij} \mid \mathcal{D}_{\mathbf{z}_{ij}}). \tag{4}$$

The latter transformation follows from the mutual independence of the edges of kinematic trees.

An important insight in our work is that the kinematic link models representing the edges can be estimated independently from the actual structure of the kinematic tree. As a result, the problem can be solved efficiently: first, we estimate the link models of all possibles edges $(ij) \in V_G \times V_G$:

$$(\hat{\mathcal{M}}_{ij}, \hat{\boldsymbol{\theta}}_{ij}) = \arg\max_{\mathcal{M}_{ij}, \boldsymbol{\theta}_{ij}} p(\mathcal{M}_{ij}, \boldsymbol{\theta}_{ij} \mid \mathcal{D}_{\mathbf{z}_{ij}}). \tag{5}$$

These link models are independent of each other and independent of whether they are actually part of the kinematic structure $E_G$. Second, given these link models, we estimate the kinematic structure. This two-step process is also visualized in Fig. 2.

Solving (5) is still a two-step process (MacKay, 2003): at the first level of inference, we assume that a particular model (e.g., like the revolute model) is true, and estimate its parameters. By applying Bayes' rule, we may write

$$\hat{\boldsymbol{\theta}}_{ij} = \arg\max_{\boldsymbol{\theta}_{ij}} p(\boldsymbol{\theta}_{ij} \mid \mathcal{D}_{\mathbf{z}_{ij}}, \mathcal{M}_{ij}) \tag{6}$$

$$= \arg\max_{\boldsymbol{\theta}_{ij}} \frac{p(\mathcal{D}_{\mathbf{z}_{ij}} \mid \boldsymbol{\theta}_{ij}, \mathcal{M}_{ij})\, p(\boldsymbol{\theta}_{ij} \mid \mathcal{M}_{ij})}{p(\mathcal{D}_{\mathbf{z}_{ij}} \mid \mathcal{M}_{ij})}. \tag{7}$$

The term $p(\boldsymbol{\theta}_{ij} \mid \mathcal{M}_{ij})$ defines the model-dependent prior over the parameter space, that we assume in our work to be uniform, and thus may be dropped. Further, we can ignore the normalizing constant $p(\mathcal{D}_{\mathbf{z}_{ij}} \mid \mathcal{M}_{ij})$, as it is has no influence on the choice of the parameter vector. This results in

$$\hat{\boldsymbol{\theta}}_{ij} = \arg\max_{\boldsymbol{\theta}_{ij}} p(\mathcal{D}_{\mathbf{z}_{ij}} \mid \boldsymbol{\theta}_{ij}, \mathcal{M}_{ij}), \tag{8}$$

which means that fitting of a link model to the observations corresponds to the problem of maximizing the data likelihood.

At the second level of inference, we need to compare the probability of different models given the data and select the model with the highest posterior:

$$\hat{\mathcal{M}}_{ij} = \arg\max_{\mathcal{M}_{ij}} \int p(\mathcal{M}_{ij}, \boldsymbol{\theta}_{ij} \mid \mathcal{D}_{\mathbf{z}_{ij}}) \, d\boldsymbol{\theta}_{ij}. \tag{9}$$

Computing the exact posterior probability of a model is in general difficult, and therefore we use in our work the Bayesian information criterion (BIC) for selecting the best model according to (9).

As a result of this inference, we obtain for each edge $(ij) \in V_G \times V_G$ a model $\hat{\mathcal{M}}_{ij}$ with parameter vector $\hat{\boldsymbol{\theta}}_{ij}$, that best describes the motions in $\mathcal{D}_{\mathbf{z}_{ij}}$ observed between these two parts. We denote this set of all possible link models with

$$\hat{\mathbb{M}} = \{(\hat{\mathcal{M}}_{ij}, \hat{\boldsymbol{\theta}}_{ij}) \mid (ij) \in V_G \times V_G\}. \tag{10}$$

Given these maximum-likelihood estimate for all links, we can now efficiently estimate the kinematic structure $E_G \subset V_G \times V_G$. For this, we aim at finding the subset, that maximizes the posterior probability of the resulting kinematic graph, i.e.,

$$\hat{E}_G = \arg\max_{E_G} \int p(E_G, \mathbb{M} \mid \mathcal{D}_{\mathbf{z}}) \, d\mathbb{M}. \tag{11}$$

We solve the equation again by maximizing the BIC over all possible structures $E_G$, using the maximum-likelihood estimate for $\hat{\mathbb{M}}$ for approximating the integral.

With this, we provide an efficient way to solve (2), by first fitting all models to the data, then selecting the best model for each link, and finally estimating the kinematic structure of the whole articulated object. From Section 2.3 to Section 2.7, we will show how to solve the model fitting problem of (8) and the model selection problem of (9) efficiently and robustly from noisy observations. In Section 2.8, we will then show how one can efficiently solve (11), given the link models. In Section 3.2, we will show how this solution for kinematic trees can be generalized to general kinematic graphs, including kinematic structures containing loops.

### 2.3 Observation Model

In the beginning, we consider simple objects consisting of only $p = 2$ rigid parts, and drop the $ij$ indices to increase readability. We consider the case that the robot has observed a sequence of $n$ relative transformations $\mathcal{D}_{\mathbf{z}} = (\mathbf{z}^1, \dots, \mathbf{z}^n)$ between two adjacent rigid parts of an articulated object. We assume the presence of Gaussian noise in each of the measurements $\mathbf{z}^n$ with zero mean and covariance $\Sigma_{\mathbf{z}} \in \mathbb{R}^{6\times 6}$.

Further, we assume that a small fraction of the observations are real outliers that cannot be explained by the Gaussian noise assumption alone. These outliers may be the result of poor perception, bad data association, or other sensor failures that are hard to be modeled explicitly. As these outliers are not related to the true value of $\boldsymbol{\Delta} = \mathbf{x}_1 \ominus \mathbf{x}_2$ at all, we assume that they come from a uniform prior, i.e., we assume a constant likelihood $p(\mathbf{z}^{\text{outlier}}) = \textit{const.}$ One can think of this as a latent variable $v \in \{0, 1\}$ indicating whether an observation is an inlier ($v = 1$) or an outlier ($v = 0$). Further, we denote with $\gamma$ the probability of drawing an outlier, i.e., $p(v = 0) = \gamma$. Our full observation model then becomes

$$\mathbf{z} \sim \begin{cases} \boldsymbol{\Delta} + \mathcal{N}(0, \Sigma_{\mathbf{z}}) & \text{if } v = 1 \\ \mathcal{U} & \text{if } v = 0 \end{cases} . \tag{12}$$

The resulting data likelihood for a single observation $\mathbf{z}$ thus is a mixture of a Gaussian and a uniform distribution with mixing constant $\gamma$:

$$p(\mathbf{z} \mid \boldsymbol{\Delta}, \gamma) = (1 - \gamma) p(\mathbf{z} \mid v = 1) + \gamma p(\mathbf{z} \mid v = 0). \tag{13}$$

Note that in general neither the true transformation $\boldsymbol{\Delta}$ nor the outlier ratio $\gamma$ are directly observable, and thus need to be estimated from the data. For comparing models with different outlier ratios, we assume a global prior of $p(\gamma) \propto \exp(-w\gamma)$ with $w$ being a weighting constant, and thereby favor models with fewer outliers over models with more outliers. The resulting data likelihood of an observation $\mathbf{z}$ given its true value $\boldsymbol{\Delta}$ thus becomes:

$$p(\mathbf{z} \mid \boldsymbol{\Delta}) = p(\mathbf{z} \mid \boldsymbol{\Delta}, \gamma) p(\gamma). \tag{14}$$

### 2.4 Candidate Models

When considering the set of objects relevant for a service robot, one quickly realizes that the joints in many objects belong to a few generic model classes. In particular, revolute and prismatic joints are used most often, although a few objects are composed of other mechanical linkages, for example spherical joints, screws, or two-bar links. Examples of revolute joints include doors, door handles, and windows. This also includes the doors of dishwashers, microwave ovens or washing machines. Examples of articulated objects belonging to the prismatic class include drawers, sliding doors, and window blinds. However, there are also objects that have different mechanical linkages, such as garage doors or two-bar office lamps. This motivates the use of a set of candidate models, that are well suited for describing the kinematic properties of a particular class of articulated links. Our candidate set consists of parametrized and non-parametrized models, in particular, it includes a model for revolute joints ($\mathcal{M}^{\text{revolute}}$), for prismatic joints ($\mathcal{M}^{\text{prismatic}}$), and rigid transformations ($\mathcal{M}^{\text{rigid}}$). Additionally, there may be articulations that do not correspond to these standard motions, for which we consider a parameter-free model ($\mathcal{M}^{\text{GP}}$). We model such joints using a combination of dimensionality reduction and Gaussian process regression.

In our framework, a model class defines the conditional probability distribution $p(\boldsymbol{\Delta} \mid \mathbf{q}, \mathcal{M}, \boldsymbol{\theta})$ and $p(\mathbf{q} \mid \boldsymbol{\Delta}, \mathcal{M}, \boldsymbol{\theta})$ by means of a *forward kinematic function* $f_{\mathcal{M},\boldsymbol{\theta}}(\mathbf{q}) = \boldsymbol{\Delta}$ and the *inverse kinematic function* $f^{-1}_{\mathcal{M},\boldsymbol{\theta}}(\mathbf{z}) = \mathbf{q}$. This means that we assume that our link models are deterministic, and we attribute all noise to measurement noise in the observations of the object parts, i.e., by means of the observation model $p(\boldsymbol{\Delta} \mid \mathbf{z})$ defined in Section 2.3.

Since we have no prior information about the nature of the connection between the two rigid parts, we do not aim to fit only a single model, but instead aim at fitting all of the candidate models to the observed data, and then we select the best model from this set.

| candidate model $\mathcal{M}$ | DOFs $d$ | parameters $k$ |
|---|---|---|
| rigid model | 0 | 6 |
| prismatic model | 1 | 9 |
| revolute model | 1 | 12 |
| Gaussian process model | $1, \ldots, 5$ | $1 + d + 6n$ |

Table 1: Overview over the different candidate models for articulated links.

### 2.5 Model Fitting using Maximum Likelihood Consensus

For estimating the parameters of any of the above-mentioned models, we need to find a parameter vector $\boldsymbol{\theta} \in \mathbb{R}^k$ that maximizes the data likelihood given the model, i.e.,

$$\hat{\boldsymbol{\theta}} = \arg\max_{\boldsymbol{\theta}} p(\mathcal{D}_{\mathbf{z}} \mid \mathcal{M}, \boldsymbol{\theta}). \tag{15}$$

In the presence of noise and outliers, finding the right parameter vector $\boldsymbol{\theta}$ that minimizes (15) is not trivial, as least squares estimation is sensitive to outliers and thus not sufficient given our observation model. Therefore, we use the MLESAC (maximum likelihood consensus) algorithm as introduced by Torr and Zisserman (2000). We estimate the initial kinematic parameters from a minimal set of randomly drawn samples from the observation sequence that we then refine using non-linear optimization of the data likelihood.

The MLESAC procedure for a model $\mathcal{M}$ works as follows: First, we generate a guess for the parameter vector $\boldsymbol{\theta}$ in (15) from a minimal set of samples from $\mathcal{D}_{\mathbf{z}}$. For this guess, we then compute the data likelihood of the whole observation sequence $\mathcal{D}_{\mathbf{z}}$ as the product over all data

$$p(\mathcal{D}_{\mathbf{z}} \mid \mathcal{M}, \hat{\boldsymbol{\theta}}) = \prod_{t=1}^{n} p(\mathbf{z}^t \mid \mathcal{M}, \hat{\boldsymbol{\theta}}). \tag{16}$$

We repeat this sampling step for a fixed number of iterations, and finally select the parameter vector maximizing (16). On this initial guess, we apply non-linear optimization on the data likelihood to refine the parameter vector using Broyden-Fletcher-Goldfarb-Shanno (BFGS) optimization, which is a quasi-Newton method for function maximization. During the maximization of the data likelihood, MLESAC iteratively also estimates the outlier ratio $\gamma$, using the Expectation Maximization algorithm.

In the following, we show for each of our link models how to (i) estimate the parameter vector $\boldsymbol{\theta}$ from a minimal sample set of observations, (ii) estimate a transformation $\mathbf{z}$ given a configuration $\mathbf{q}$, and (iii) estimate the configuration $\mathbf{q}$ given a transformation $\mathbf{z}$. A brief overview over all model candidates is given in Table 1.

#### 2.5.1 Rigid Model

We parametrize a rigid link by a fixed relative transformation between two object parts. Thus, the parameter vector $\boldsymbol{\theta}$ has $k = 6$ dimensions. During the sampling consensus step, we draw a single observation $\mathbf{z}$ from the training data $\mathcal{D}_{\mathbf{z}}$ that gives us an initial guess for the parameter vector $\hat{\boldsymbol{\theta}}$. This parameter vector thus corresponds to the estimated fixed relative transformation between the two parts. For the rigid transformation model, the forward kinematics function equals the parameter vector as it corresponds to the estimated fixed relative transform between the two parts:

$$f_{\mathcal{M}^{\text{rigid}},\boldsymbol{\theta}}(\mathbf{q}) = \boldsymbol{\theta}. \tag{17}$$

As the rigid model has zero DOFs ($d = 0$), an inverse kinematic function is not needed.

### 2.5.2 Prismatic Model

Prismatic joints move along a single axis, and thus have a one-dimensional configuration space. The prismatic model describes a translation along a vector of unit length $\mathbf{e} \in \mathbb{R}^3$ relative to some fixed origin $\mathbf{a} \in SE(3)$. This results in a parameter vector $\boldsymbol{\theta} = (\mathbf{a}; \mathbf{e})$ with $k = 9$ dimensions.

For estimating these parameters, we sample two observations from the training data. For this, we pick the transformation of the first sample as the origin $\mathbf{a}$ and the normalized vector between them as the prismatic axis $\mathbf{e}$.

A configuration $q \in \mathbb{R}$ then encodes the distance from the origin $\mathbf{a}$ along the direction of motion $\mathbf{e}$. The forward kinematics function for the prismatic model $\mathcal{M}^{\text{prismatic}}$ is

$$f_{\mathcal{M}^{\text{prismatic}},\boldsymbol{\theta}}(\mathbf{q}) = \mathbf{a} \oplus q\mathbf{e}. \tag{18}$$

Let $trans(\cdot)$ be the function that removes all rotational components. The inverse kinematic function then becomes

$$f^{-1}_{\mathcal{M}^{\text{prismatic}},\boldsymbol{\theta}}(\mathbf{z}) =< \mathbf{e}, trans(\mathbf{a} \ominus \mathbf{z}) >, \tag{19}$$

where $< \cdot, \cdot >$ refers to the dot product.

### 2.5.3 Revolute Model

The revolute model describes the motion of a revolute joint, i.e., a one-dimensional motion along a circular arc. We parametrize this model by the center of rotation $\mathbf{c} \in SE(3)$, and a rigid transformation $\mathbf{r} \in SE(3)$, from the center to the moving part. This yields a parameter vector $\boldsymbol{\theta} = (\mathbf{c}; \mathbf{r})$ with $k = 12$ dimensions.

For the revolute model, we sample three observations $\mathbf{z}^i$, $\mathbf{z}^j$ and $\mathbf{z}^k$ from the training data. First, we estimate the plane spanned by these three points; its plane normal then is parallel to the rotation axis. Second, we compute the circle center as the intersection of the perpendicular lines of the line segments between the three observations. Together with the rotation axis, this gives us the center of rotation $\mathbf{c}$. Finally, we estimate the rigid transformation $\mathbf{r}$ of the circle from the first sample.

For the forward kinematic function, we obtain for revolute links

$$f_{\mathcal{M}^{\text{revolute}},\boldsymbol{\theta}}(\mathbf{q}) = \mathbf{c} \oplus Rot_Z(\mathbf{q}) \oplus \mathbf{r}, \tag{20}$$

where $Rot_Z(\mathbf{q})$ denotes a rotation around the Z-axis by $\mathbf{q}$. Thus, $\mathbf{q} \in \mathbb{R}$ specifies the angle of rotation. For estimating the configuration of a revolute joint we use

$$f^{-1}_{\mathcal{M}^{\text{revolute}},\boldsymbol{\theta}}(\mathbf{z}) = Rot_Z^{-1}((\mathbf{z} \ominus \mathbf{c}) - \mathbf{r}), \tag{21}$$

where $Rot_Z^{-1}(\cdot)$ gives the rotation around the Z-axis.

### 2.5.4 Gaussian Process Model

Although rigid transformations in combination with revolute and prismatic joints might seem at the first glance to be sufficient for a huge class of kinematic objects, many real-world objects cannot be described by a single shifting or rotation axis. Examples for such objects include garage doors or office table lamps, but also furniture whose joints have aged and became loose.

Therefore, we provide additionally a non-parametric model which is able to describe more general kinematic links. This model is based on dimensionality reduction for discovering the latent manifold of the configuration space and Gaussian process regression for learning a generative model. Consider the manifold that is described by the observations $\mathcal{D}_{\mathbf{z}}$ between two rigid bodies. Depending on the number of DOFs $d$ of this link, the data samples will lie on or close to a $d$-dimensional manifold with $1 \leq d \leq 6$ being non-linearly embedded in $SE(3)$.

There are many different dimensionality reduction techniques such as principal component analysis (PCA) for linear manifolds, or Isomap and locally linear embedding (LLE) for non-linear manifolds (Tenenbaum, de Silva, & Langford, 2000; Roweis & Saul, 2000). In our experiments, we used both PCA and LLE for dimensionality reduction. PCA has the advantage of being more robust against noise for near-linear manifolds, while LLE is more general and can also model strongly non-linear manifolds.

The general idea here is that we use the dimensionality reduction technique to obtain the inverse kinematics function $f^{-1}_{\mathcal{M}^{\text{GP}}} : SE(3) \to \mathbb{R}^d$. As a result, we can assign configurations to each of the observations, i.e.,

$$f^{-1}_{\mathcal{M}^{\text{GP}}}(\mathbf{z}) = \mathbf{q}. \tag{22}$$

These assignments of observations to configurations can now be used to learn the forward kinematics function $f_{\mathcal{M}^{\text{GP}},\boldsymbol{\theta}}(\cdot)$ from the observations. Except for linear actuators, we expect this function to be strongly non-linear.

A flexible approach for solving such non-linear regression problems given noisy observations are Gaussian processes (GPs). One of the main features of the Gaussian process framework is that the observed data points are explicitly included in the model. Therefore, no parametric form of $f_{\mathcal{M}^{\text{GP}}} : \mathbb{R}^d \to SE(3)$ needs to be specified. Data points can be added to a GP at any time, which facilitates incremental and online learning. For this model, we aim to learn a GP that fits the dependency

$$f_{\mathcal{M}^{\text{GP}}}(\mathbf{q}) + \boldsymbol{\epsilon} = \mathbf{z} \tag{23}$$

for the unknown forward model underlying the articulated link under consideration. We assume homoscedastic noise, i.e., independent and identically, normally distributed noise terms $\boldsymbol{\epsilon} \sim \mathcal{N}(0, \Sigma_{\mathbf{z}})$. For simplicity, we train 12 independent Gaussian processes for the free components of a homogeneous $4 \times 4$ transformation matrix. As a consequence of this over-parametrization, the predicted transformation matrices are not necessarily valid. In practice, however, they are very close to valid transformation matrices, that can be found using ortho-normalization via singular value decomposition. In our approach, we use the standard choice for the covariance function, the squared exponential. It describes the relationship between two configurations $\mathbf{q}_i$ and $\mathbf{q}_j$ in configuration space by

$$k(\mathbf{q}_i, \mathbf{q}_j) = \sigma_f^2 \exp\left(-\frac{1}{2}(\mathbf{q}_i - \mathbf{q}_j)^T \Lambda^{-1} (\mathbf{q}_i - \mathbf{q}_j)\right), \tag{24}$$

where $\sigma_f^2$ is the signal variance, and $\Lambda^{-1} = \mathrm{diag}(l_1, \ldots, l_d)$ is the diagonal matrix of the length-scale parameters. This results in a $(1 + d)$-dimensional hyper-parameter vector $\boldsymbol{\theta} = (\sigma_f^2, l_1, \ldots, l_d)$. As GPs are data-driven, they require all training data when making predictions. Therefore, we count all data samples as parameters of our model, so that the number of parameters becomes $k = (1 + d) + 6n$, where $n = |\mathcal{D}_\mathbf{z}|$ is the number of observations. We refer the interested reader to the text book by Rasmussen and Williams (2006) for more details about GP regression.

Note that this GP link model directly generalizes to higher-dimensional configuration spaces, i.e., with $d > 1$: after the dimensionality reduction from observations in $SE(3)$ to configurations in $\mathbb{R}^d$, we can again learn a Gaussian process regression that learns the mapping from the configuration space $\mathbb{R}^d$ back to transformations in $SE(3)$. Note that this GP model that we present here is similar to the GPLVM model introduced by Lawrence (2005). In contrast to GPLVM, we do not optimize the latent configurations for maximizing the data likelihood. This would invalidate our inverse kinematics function (22), and limits the GPLVM model to map only from latent space to data space. With our approach, we can also infer the configuration of new relative transformations not available during training.

### 2.6 Model Evaluation

To evaluate how well a single observation $\mathbf{z}$ is explained by a model, we have to evaluate $p(\mathbf{z} \mid \mathcal{M}, \boldsymbol{\theta})$. As the configuration is latent, i.e., not directly observable by the robot, we have to integrate over all possible values of $\mathbf{q}$, i.e.,

$$p(\mathbf{z} \mid \mathcal{M}, \boldsymbol{\theta}) = \int p(\mathbf{z} \mid \mathbf{q}, \mathcal{M}, \boldsymbol{\theta}) p(\mathbf{q} \mid \mathcal{M}, \boldsymbol{\theta}) \, d\mathbf{q}. \tag{25}$$

Under the assumption that all DOFs of the link are independent of each other, and that no configuration state $\mathbf{q}$ is more likely than another (or equivalently, that $p(\mathbf{q} \mid \mathcal{M}, \boldsymbol{\theta})$ is uniformly distributed), we may write

$$p(\mathbf{q} \mid \mathcal{M}, \boldsymbol{\theta}) \approx n^{-d}, \tag{26}$$

where $n = |\mathcal{D}_\mathbf{z}|$ is the number of observations so far, and thus the number of estimated configurations in the $d$-dimensional configuration space. With this, (25) can be simplified to

$$p(\mathbf{z} \mid \mathcal{M}, \boldsymbol{\theta}) \approx n^{-d} \int p(\mathbf{z} \mid \mathbf{q}, \mathcal{M}, \boldsymbol{\theta}) \, d\mathbf{q}. \tag{27}$$

If we assume that $p(\mathbf{z} \mid \mathbf{q}, \mathcal{M}, \boldsymbol{\theta})$ is an uni-modal distribution, an approximation of the integral is to evaluate it only at the estimated configuration $\hat{\mathbf{q}}$ given the observation $\mathbf{z}$ using the inverse kinematics function of the model under consideration, i.e.,

$$\hat{\mathbf{q}} = f^{-1}_{\mathcal{M},\boldsymbol{\theta}}(\mathbf{z}). \tag{28}$$

For this configuration, we can compute the expected transformation $\hat{\boldsymbol{\Delta}}$ using the forward kinematics function of the model,

$$\hat{\boldsymbol{\Delta}} = f_{\mathcal{M},\boldsymbol{\theta}}(\hat{\mathbf{q}}). \tag{29}$$

Given the observation $\mathbf{z}$ and the expected transformation $\hat{\boldsymbol{\Delta}}$, we can now efficiently compute the data likelihood of (27) using the observation model from (14) as

$$p(\mathbf{z} \mid \mathcal{M}, \boldsymbol{\theta}) \approx n^{-d} p(\mathbf{z} \mid \hat{\boldsymbol{\Delta}}). \tag{30}$$

Note that the approximation of the integral based on the forward and inverse kinematics model corresponds to a projection of the noisy observations onto the model. Finally, the marginal data likelihood over the whole observation sequence becomes

$$p(\mathcal{D}_{\mathbf{z}} \mid \mathcal{M}, \boldsymbol{\theta}) = \prod_{\mathbf{z} \in \mathcal{D}_{\mathbf{z}}} p(\mathbf{z} \mid \mathcal{M}, \boldsymbol{\theta}). \tag{31}$$

### 2.7 Model Selection

After having fitted all model candidates to an observation sequence $\mathcal{D}_{\mathbf{z}}$, we need to select the model that explains the data best. For Bayesian model selection, this means that we need to compare the posterior probability of the models given the data

$$p(\mathcal{M} \mid \mathcal{D}_{\mathbf{z}}) = \int \frac{p(\mathcal{D}_{\mathbf{z}} \mid \mathcal{M}, \boldsymbol{\theta}) p(\boldsymbol{\theta} \mid \mathcal{M}) p(\mathcal{M})}{p(\mathcal{D}_{\mathbf{z}})} \, d\boldsymbol{\theta}. \tag{32}$$

While the evaluation of the model posterior is in general difficult, it can be approximated efficiently based on the Bayesian information criterion (BIC) (Schwarz, 1978). We denote with $k$ the number of parameters of the current model under consideration, and $n$ the number of observations in the training data. Then, the BIC is defined as

$$\mathrm{BIC}(\mathcal{M}) = -2 \log p(\mathcal{D}_{\mathbf{z}} \mid \mathcal{M}, \hat{\boldsymbol{\theta}}) + k \log n, \tag{33}$$

where $\hat{\boldsymbol{\theta}}$ is the maximum likelihood parameter vector. Model selection now reduces to selecting the model that has the lowest BIC, i.e.,

$$\hat{\mathcal{M}} = \arg\min_{\mathcal{M}} \mathrm{BIC}(\mathcal{M}). \tag{34}$$

We refer the interested reader to the work of Bishop (2007) for further information on the BIC.

### 2.8 Finding the Connectivity

So far, we ignored the question of connectivity and described how to evaluate and select a model $\mathcal{M}$ for a single link between two parts of an object only. In this section, we extend our approach to efficiently find kinematic trees for articulated objects consisting of multiple parts.

We adopt the connectivity model from Featherstone and Orin (2008) for modeling the kinematic structure as an undirected graph $G = (V_G, E_G)$. The nodes $V_G$ in this graph correspond to the poses of the individual object parts, while the edges $E_G$ correspond to the links between these parts. We will now re-introduce the $ij$-indices, i.e., use $\mathcal{D}_{\mathbf{z}_{ij}}$ to refer to the observations of link $(ij)$, and $\mathcal{D}_{\mathbf{z}}$ to refer to the observations of the whole articulated object. $\mathcal{D}_{\mathbf{z}}$ thus contains the observations of all edges in the graph $G$, i.e., $\mathcal{D}_{\mathbf{z}} = \{\mathcal{D}_{\mathbf{z}_{ij}} \mid (ij) \in E_G\}$. In the previous section, we established an algorithm that fits and selects for any given edge $(ij)$ in this graph a corresponding link model $\hat{\mathcal{M}}_{ij}$ with parameter vector $\hat{\boldsymbol{\theta}}_{ij}$. Given this, we now need to select the kinematic structure $E_G$, i.e., which of these link models are actually present in the articulated object under consideration.

For the moment, we will consider only kinematic tree mechanisms, i.e., mechanisms without kinematic loops. Now, we consider a fully connected graph with $p$ vertices, i.e., one vertex for each object part of the articulated object. The set of possible kinematic trees for the articulated object is now given by all spanning trees of this graph. The endeavor of explicitly computing, evaluating, and reasoning with all kinematic trees, however, is not tractable in practice.

We therefore seek to find the kinematic structure $E_G$ that maximizes the posterior as stated previously in (11),

$$\hat{E}_G = \arg\max_{E_G} p(E_G \mid \mathcal{D}_{\mathbf{z}}) \tag{35}$$

$$= \arg\max_{E_G} p(\{(\hat{\mathcal{M}}_{ij}, \hat{\boldsymbol{\theta}}_{ij}) \mid (ij) \in E_G\} \mid \mathcal{D}_{\mathbf{z}}) \tag{36}$$

$$= \arg\max_{E_G} \prod_{(ij)\in E_G} p(\hat{\mathcal{M}}_{ij}, \hat{\boldsymbol{\theta}}_{ij} \mid \mathcal{D}_{\mathbf{z}}) \tag{37}$$

$$= \arg\max_{E_G} \sum_{(ij)\in E_G} \log p(\hat{\mathcal{M}}_{ij}, \hat{\boldsymbol{\theta}}_{ij} \mid \mathcal{D}_{\mathbf{z}}). \tag{38}$$

Note that because of the independence assumption of the individual links for kinematic trees, the posterior of the kinematic model for the whole object in (36) can be written as the product over the posteriors of the individual links in (37). After taking the logarithm in (38), the structure selection problem takes a form that can be solved efficiently. The key insight here is that the kinematic tree that maximizes (38) corresponds to the problem of selecting the minimum spanning tree in a fully connected graph with edge costs corresponding to the negative log posterior,

$$\text{cost}_{ij} = -\log p(\mathcal{M}_{ij}, \boldsymbol{\theta}_{ij} | \mathcal{D}_{\mathbf{z}_{ij}}), \tag{39}$$

that we approximate with the BIC value. The sum over these edge costs then corresponds to the negative log posterior of the kinematic tree, and a minimum spanning tree thus maximizes the posterior of (38). The best kinematic structure can now be found efficiently, i.e., in $\mathcal{O}(p^2 \log p)$ time, using for example Prim's or Kruskal's algorithm for finding minimum spanning trees (Cormen, Leiserson, Rivest, & Stein, 2001).

## 3. Framework Extensions

The approach described so far enables a robot to learn kinematic models of articulated objects from scratch. In the following, we will consider three extensions. The first extension enables a robot to exploit priors learned from previous interactions when learning new models. Second, we generalize our framework to general kinematic graphs, i.e., consider additionally objects that contain closed kinematic chains. Third, we show that estimating the number of DOFs of articulated objects follows directly from our approach.

### 3.1 Learning and Exploiting Priors

Using the approach described above, a robot always starts learning a model from scratch when it observes movements of a new articulated object. From a learning perspective, this may be seen as unsatisfactory since most articulated objects encountered in man-made environments belong to few different classes with similar parameters. For example, in a specific office or kitchen, many cabinet doors will open in the same way, i.e., have the same radius and rotation axis. Furthermore, in some countries the size of such furniture is standardized. Thus, a robot operating in such environments over extended periods of time can significantly boost its performance by learning priors over the space of possible articulated object models.

This section describes our approach for learning priors for articulated objects and a means for exploiting them as early as possible while manipulating a previously unseen articulated object. In addition to the previous section, we explicitly want to transfer model information contained in already learned models to newly seen articulated objects. The key idea here is to identify a small set of representative models for the articulated objects and to utilize this as prior information to increase the prediction accuracy when handling new objects.

To keep the notation simple, consider the case that we have previously encountered two articulated objects consisting of two parts and thus a single link only. Their observed motion is given by two observation sequences $\mathcal{D}_{\mathbf{z},1}$ and $\mathcal{D}_{\mathbf{z},2}$. The question now is whether both trajectories should be described by two distinct models $\mathcal{M}_1$ and $\mathcal{M}_2$ or by a joint model $\mathcal{M}_{1+2}$. In the first case, we can split the posterior as the two models are mutually independent, i.e.,

$$p(\mathcal{M}_1, \mathcal{M}_2 \mid \mathcal{D}_{\mathbf{z},1}, \mathcal{D}_{\mathbf{z},2}) = p(\mathcal{M}_1 \mid \mathcal{D}_{\mathbf{z},1}) p(\mathcal{M}_2 \mid \mathcal{D}_{\mathbf{z},2}). \tag{40}$$

In the latter case, both trajectories are explained by a single, joint model $\mathcal{M}_{1+2}$ with a parameter vector $\boldsymbol{\theta}_{1+2}$, that is estimated from the joint data $\mathcal{D}_{\mathbf{z},1} \cup \mathcal{D}_{\mathbf{z},2}$. For future reference, we denote the corresponding posterior probability as

$$p(\mathcal{M}_{1+2} \mid \mathcal{D}_{\mathbf{z},1}, \mathcal{D}_{\mathbf{z},2}). \tag{41}$$

We can determine whether a joint model is better than two separate models by comparing the posterior probabilities from (40) and (41), i.e, by evaluating

$$p(\mathcal{M}_{1+2} \mid \mathcal{D}_{\mathbf{z},1}, \mathcal{D}_{\mathbf{z},2}) > p(\mathcal{M}_1 \mid \mathcal{D}_{\mathbf{z},1}) p(\mathcal{M}_2 \mid \mathcal{D}_{\mathbf{z},2}). \tag{42}$$

This expression can be efficiently evaluated by using the BIC as follows. The joint model is learned from $n = n_1 + n_2$ data points, using $k$ parameters, with a data likelihood of $L = p(\mathcal{M}_{1+2} \mid \mathcal{D}_{\mathbf{z},1}, \mathcal{D}_{\mathbf{z},2})$, while the two separate models are learned from $n_1$ and $n_2$ samples, using $k_1$ and $k_2$ parameters, and data likelihoods of $L_1 = p(\mathcal{D}_{\mathbf{z},1} \mid \mathcal{M}_1)$ and $L_2 = p(\mathcal{D}_{\mathbf{z},2} \mid \mathcal{M}_2)$, respectively. Accordingly, we check whether

$$\mathrm{BIC}(\mathcal{M}_{1+2} \mid \mathcal{D}_{\mathbf{z},1}, \mathcal{D}_{\mathbf{z},2}) < \mathrm{BIC}(\mathcal{M}_1 \mid \mathcal{D}_{\mathbf{z},1}) + \mathrm{BIC}(\mathcal{M}_2 \mid \mathcal{D}_{\mathbf{z},2}) \tag{43}$$

i.e., whether

$$-2 \log L + k \log n < -2 \log(L_1 L_2) + k_1 \log n_1 + k_2 \log n_2. \tag{44}$$

Informally, merging two models into one is beneficial if the joint model can explain the data equally well (i.e., $L \approx L_1 L_2$), while requiring only a single set of parameters.

If more than two trajectories are considered, one has to evaluate all possible assignments of these trajectories to models and select the assignment with the highest posterior. As this quickly becomes intractable due to combinatorial explosion, we use an approximation and consider the trajectories sequentially and in the order the robot observes them. We check whether merging the new trajectory with one of the existing models leads to a higher posterior compared to adding a new model for that trajectory to the set of previously encountered models.

After having identified a set of models as prior information, we can exploit this knowledge for making better predictions when observing a so far unseen articulated object. Consider the situation in which a partial trajectory of a new object has been observed. To exploit the prior information, we proceed exactly as before. We compute and compare the posteriors according to (44), treating the newly observed data points as a new model or respectively merging them into one of the $w$ previously identified models by evaluating

$$p(\mathcal{M}_{new}, \mathcal{M}_1, \ldots, \mathcal{M}_w) < \max_{j=1,\ldots,w} p(\mathcal{M}_1, \ldots, \mathcal{M}_{j+new}, \ldots, \mathcal{M}_w). \tag{45}$$

If the newly observed data is merged with an existing model, the parameter vector is estimated from a much larger dataset $\mathcal{D}_{\mathbf{z},j} \cup \mathcal{D}_{\mathbf{z},new}$ instead of $\mathcal{D}_{\mathbf{z},new}$ which leads to a better estimation. Note that this step is carried out after each observation of the new sequence. Thus, if the currently manipulated object ceases to be explained by the known models, the method instantaneously creates a new model. After successful object manipulation, this model serves as additional prior information for the future.

### 3.2 Closed Kinematic Chains

Although most articulated objects have the connectivity of kinematic trees, there exist mechanisms containing closed kinematic chains (Featherstone & Orin, 2008). An intuitive example of a closed-loop system is a robot that opens a door with its manipulator. While both the robot and the door can be described individually as a kinematic tree using our approach, the combined system of the robot, the door and the floor creates a kinematic loop. Another example is a humanoid robot that has multiple contact points, e.g., by standing on both feet, or a robot that manipulates an object with two arms (Sentis, Park, & Khatib, 2010). To describe such closed-loop systems, we need to extend our approach.

Recall that for finding the kinematic structure in Section 2.8, we established the correspondence for finding the graph that maximizes the posterior probability. For that, we needed to compute the data likelihood of the graph based on edge constraints, which was easy for kinematic trees. In this case, all links can be evaluated independently of each other. However, computing the data likelihood of a kinematic graph based on edge constraints is more difficult. This results from the more complex (joint) prediction of the poses of the object parts involved in such a kinematic loop. In general, the chained up predictions of the relative transformations between the object parts will not lead to a globally consistent prediction, which is needed to compute the overall data likelihood in case of closed kinematic chains.

This problem, however, is closely related to loop-closing in the graph-based formulation of the simultaneous localization and mapping (SLAM) problem (Lu & Milios, 1997; Dellaert, 2005; Frese, 2006; Grisetti, Stachniss, & Burgard, 2009). For this type of problem, closed-form solutions exist only for very simple cases. A popular solution for the general case are iterative optimization approaches to deal with the underlying non-linear least squares problem.

To obtain a consistent pose estimation for the whole graph, we use the optimization engine HOG-Man by Grisetti, Kümmerle, Stachniss, Frese, and Hertzberg (2010), originally designed to solve the SLAM problem. To generate the input graph for HOG-Man, we proceed as follows. We add a vertex for each object part representing its initial pose $\hat{\mathbf{x}}'_1, \ldots, \hat{\mathbf{x}}'_n$, that we estimate for an (arbitrary) spanning tree of the graph. Then, for each link model $\mathcal{M}_{ij}$ in our graph $G$, we add an edge that constrains the relative transformation between $\hat{\mathbf{x}}'_i$ and $\hat{\mathbf{x}}'_j$ to the expected transformation $\hat{\boldsymbol{\Delta}}_{ij}$ (in SLAM, this corresponds to an observation). The optimization approach will then compute a set of new poses $\hat{\mathbf{x}}_1, \ldots, \hat{\mathbf{x}}_n$ that is in line with the constraints in the sense that it is the best prediction in terms of the squared error that can be obtained given the links (in SLAM, this corresponds to the corrected trajectory of the robot).

For the pose observations $\mathbf{y}_i$, we assume Gaussian noise with zero mean and covariance $\Sigma_{\mathbf{y}}$, i.e.,

$$\mathbf{y}_i = \mathbf{x}_i + \boldsymbol{\epsilon} \tag{46}$$

$$\boldsymbol{\epsilon} \sim \mathcal{N}(0, \Sigma_{\mathbf{y}}). \tag{47}$$

The data likelihood for a single object part being observed at pose $\mathbf{y}$ while being expected at pose $\hat{\mathbf{x}}$ given the kinematic graph $G$ and a configuration $\mathbf{q}$ then becomes

$$p(\mathbf{y}_i \mid G, \mathbf{q}) \propto \exp\left(-\frac{1}{2}(\hat{\mathbf{x}}_i \ominus \mathbf{y}_i)^T \Sigma_{\mathbf{y}}^{-1} (\hat{\mathbf{x}}_i \ominus \mathbf{y}_i)\right). \tag{48}$$

Using this, the *global* data likelihood of an articulated object in a particular configuration can be computed as the product over the likelihoods of all individual object parts, i.e.,

$$p(\mathbf{y}_{1:p} \mid G, \mathbf{q}) = \prod_{i \in 1,\ldots,p} p(\mathbf{y}_i \mid G, \mathbf{q}). \tag{49}$$

As the configuration $\mathbf{q}$ of the articulated object is latent and thus not known, we need to integrate over all possible configurations, i.e., calculate

$$p(\mathbf{y}_{1:p} \mid G) = \int p(\mathbf{y}_{1:p} \mid G, \mathbf{q}) p(\mathbf{q} \mid G) d\mathbf{q}. \tag{50}$$

Similar to (25), we approximate this integral by evaluating it only at the most likely configuration $\hat{\mathbf{q}}$ of the articulated object. We assume that the configurations $\mathbf{q}$ are uniformly distributed, i.e., then $p(\mathbf{q} \mid G) \approx n^{-D}$, where $n$ is the number of pose observations and $D$ is the total number of DOFs of the articulated object. The data likelihood for a pose observation $\mathbf{y}_{1:p}$ then becomes

$$p(\mathbf{y}_{1:p} \mid G) \approx n^{-D} p(\mathbf{y}_{1:p} \mid G, \hat{\mathbf{q}}). \tag{51}$$

The data likelihood of an observation sequence $\mathcal{D}_{\mathbf{y}} = (\mathbf{y}^1_{1:p}, \ldots, \mathbf{y}^n_{1:p})$ of a whole articulated object then is

$$p(\mathcal{D}_{\mathbf{y}} \mid G) \approx \prod_{i \in 1,\ldots,n} n^{-D} p(\mathbf{y}^i_{1:p} \mid G, \hat{\mathbf{q}}^i) \tag{52}$$

$$= n^{-nD} \prod_{i \in 1,\ldots,n} p(\mathbf{y}^i_{1:p} \mid G, \hat{\mathbf{q}}^i). \tag{53}$$

This data likelihood can now be used to select the best kinematic structure. Note that in principle, all possible graphs need to be evaluated – which is super-exponential in the number of object parts, and polynomial in the number of template models. In contrast, finding the exact solution in the case of kinematic trees has a polynomial complexity of only $\mathcal{O}(mp^2)$. Obviously, the massive set of possible graph structures can only be fully evaluated for small articulated objects and few template models.

In the absence of an efficient and exact solution, we propose an efficient approximation that is able to find the locally best graph from an initial guess using a randomized search strategy in polynomial time. The idea is that given the spanning tree as an initial solution we only evaluate the graphs in the neighborhood of the current structure, i.e., graphs whose topology is similar to the current one, e.g., by adding or removing one edge at a time. As we will see in the experimental section, this heuristic is able to find the optimal (or near-optimal) graph structure in most of the cases. Additionally, we can guarantee that this randomized search strategy never gets worse than the initial solution, i.e., in our case the spanning tree.

### 3.3 Finding the Number of DOFs

The current configuration of an articulated object is given by the stacked vector of all individual configurations of its articulated links, i.e.,

$$\mathbf{q}_{\text{links}} = \begin{pmatrix} \mathbf{q}_1 \\ \mathbf{q}_2 \\ \vdots \\ \mathbf{q}_{D_{\text{links}}} \end{pmatrix} \tag{54}$$

. The question now is, whether the articulated object actually has as many DOFs as the sum of DOFs of its individual links might suggest. Clearly, in the case that the articulated object is a kinematic tree, the DOFs $D_{\text{object}}$ of the articulated object directly equals the sum over the DOFs of its links $D_{\text{links}} = \sum_{(ij) \in E_G} d_{ij}$ as all of its links can be actuated independently of each other. However, for articulated objects containing loops, finding the number of DOFs of an articulated object is not trivial.

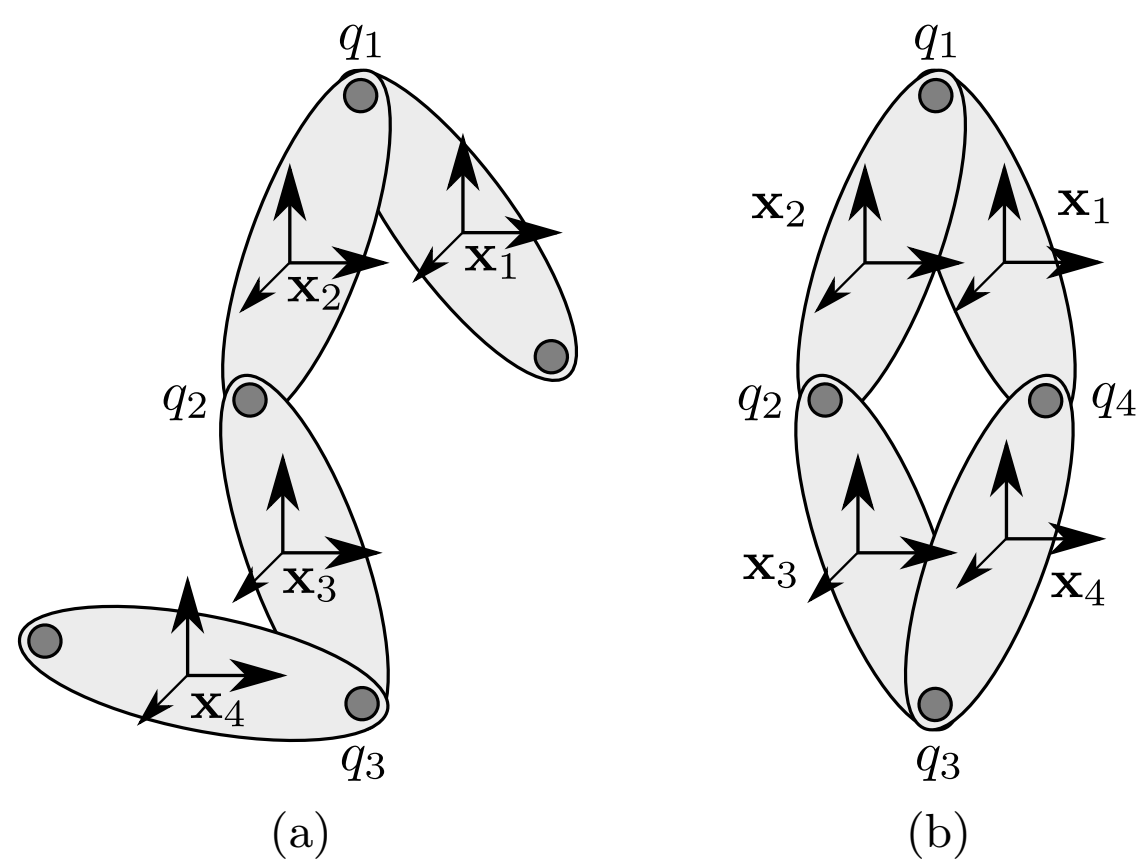

Figure 4: Example of an open and a closed kinematic chain. Whereas the open chain in (a) has three DOFs, the closed chain (b) has also only a single DOF.

For an example, consider the object in Fig. 4a which consists of four object parts and a total of three DOFs. In contrast, the object in Fig. 4b consists of four object parts, connected by four revolute links in the form of a loop. Each of the four links has a single DOF, and therefore the configuration vector defining the configuration of all of its links is $\mathbf{q}_{\text{links}} = (q_1, q_2, q_3, q_4) \in \mathbb{R}^4$. Yet, the overall system has only a single DOF: when the first joint is brought into a particular configuration, the other joints are fixed as well, as a result of the loop closure. This means that the object configuration $\mathbf{q}_{\text{object}} \in \mathbb{R}$ has only a single dimension, and thus the object configuration space is a one-dimensional manifold embedded in the four-dimensional link configuration space.

Finding a mapping between the high-dimensional link configuration space $\mathbb{R}^{D_{\text{links}}}$ and a lower-dimensional object configuration space $\mathbb{R}^{D_{\text{object}}}$ can for example be achieved using PCA for linear manifolds, and LLE or ISOMAP for non-linear manifolds. In the case of PCA, this results in finding a projection matrix $P \in \mathbb{R}^{D_{\text{object}} \times D_{\text{links}}}$ describing the mapping

$$\mathbf{q}_{\text{object}} = P\mathbf{q}_{\text{links}} \tag{55}$$

Recall from (53), that the number of DOFs has a strong influence on the data likelihood of a configuration, because a higher dimensional configuration space results in a lower likelihood for a single configuration. As a result, a model with fewer DOFs is preferred over a model with more DOFs. At the same time, if additional parameters need to be estimated for the dimension reduction, then these parameters are also model parameters and thus need to be considered during model selection.

Informally speaking, if a kinematic graph with fewer DOFs explains the data equally well, it will have a higher data likelihood and thus it be favored in the structure selection step. In the experimental section, we will see that we can use this to accurately and robustly estimate the DOFs of various articulated objects.

## 4. Perception of Articulated Objects

For estimating the kinematic model of an articulated object, our approach needs a sequence of $n$ pose observations $\mathcal{D}_{\mathbf{y}} = (\mathbf{y}^1_{1:p}, \ldots, \mathbf{y}^n_{1:p})$ that includes the poses of all $p$ parts of the object. For our experiments, we used different sources for acquiring these pose observations: marker-based perception, as described in Section 4.1, domain-specific but marker-less perception as described in Section 4.2, and perception based on the internal forward kinematic model of a manipulator using its joint encoders as described in Section 4.3.

### 4.1 Marker-Based Perception

For observing the pose of an articulated object, we used in our experiments three different marker-based systems, each with different noise and outlier characteristics: a motion capture studio with low noise and no outliers, ARToolkit markers with relatively high noise and frequent outliers, and OpenCV's checkerboard detector with moderate noise and occasional outliers.

#### 4.1.1 Motion Capturing Studio

We conducted our first experiments in a PhaseSpace motion capture studio at Willow Garage, in collaboration with Pradeep and Konolige (Sturm et al., 2009). This tracking system uses several high-speed cameras installed on a rig along the ceiling, and active LED markers attached on the individual parts of the articulated object. The data from the PhaseSpace device is virtually noise- and outlier-free. The noise of the PhaseSpace system is specified to be $\sigma_{\mathbf{y},\text{pos}} < 0.005\,\text{m}$ and $\sigma_{\mathbf{y},\text{orient}} < 1^\circ$.

#### 4.1.2 ARToolkit Markers

Additionally, we used the passive marker-based system ARToolkit for registering the 3D pose of objects by Fiala (2005). This system has the advantage that it requires only a single camera, and can be used without any further infrastructure. The ARToolkit markers consist of a black rectangle and an error-correcting code imprinted on a 6x6-grid inside the rectangle for distinguishing the individual markers. We found that the noise with this system strongly depends on the distance and the angle of the marker to the camera. With a marker size of 0.08 m and in a distance of 2 m from the camera, we typically obtained noise values of $\sigma_{\mathbf{y},\text{pos}} = 0.05\,\text{m}$ and $\sigma_{\mathbf{y},\text{orient}} = 15^\circ$.

#### 4.1.3 Checkerboard Markers

OpenCV's checkerboard detector provides a much higher pose accuracy. The detector searches the camera images for strong black and white corners at sub-pixel accuracy (Bradski & Kaehler, 2008). With this system, we typically obtained measurement noise around $\sigma_{\mathbf{y},\text{pos}} = 0.005\,\text{m}$ and $\sigma_{\mathbf{y},\text{orient}} = 5^\circ$ with marker sizes of 0.08 m side length in 2 m distance from the camera. One can distinguish different markers with this system by using checkerboards with varying numbers of rows and columns.

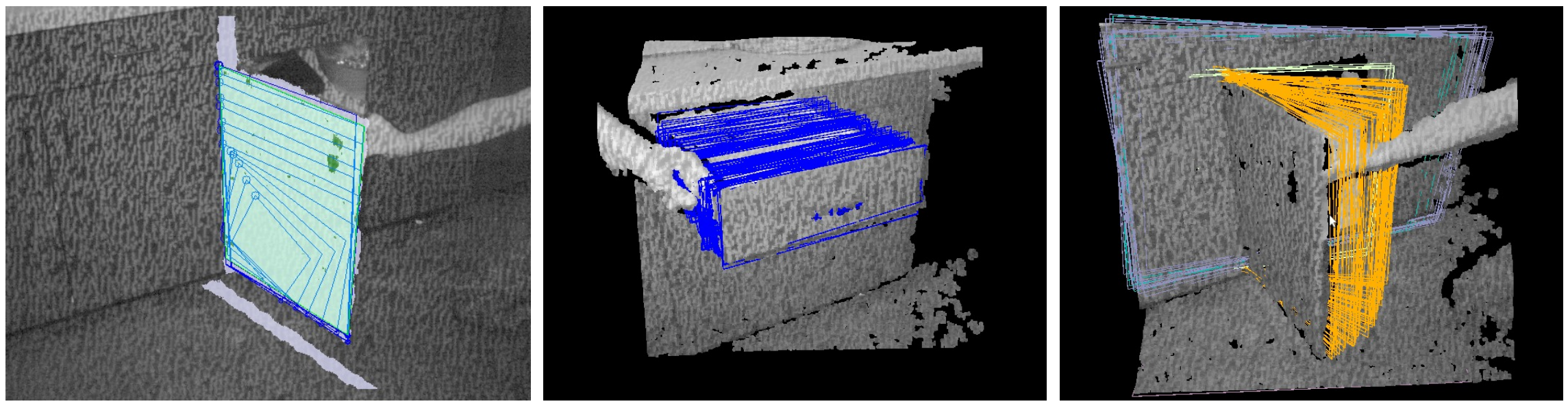

Figure 5: Marker-less pose estimation using a stereo camera. For each segmented plane, we iteratively fit a rectangle (left image). The right two images show observed tracks of a cabinet drawer and a cabinet door.

### 4.2 Marker-Less Pose Estimation

In contrast to using artificial markers, it is also possible to estimate the object pose directly, for example, from dense depth images acquired by a stereo camera system. We recently developed such a system in collaboration with Konolige (Sturm et al., 2010). Using a marker-less camera-based tracking system has several advantages. First, it does not rely on artificial markers attached to the articulated objects, and second, it does not require expensive range scanners which have the additional disadvantage that they poorly deal with moving objects, making them inconvenient for learning articulation models. However, we recognize that in general the registration of arbitrary objects in point clouds is still an open issue. Therefore, we restrict ourselves here to fronts of kitchen cabinets. This does not solve the general perception problem, but provides a useful and working solution for mobile manipulation robots performing service tasks in households. In our concrete scenario, the perception of articulated drawers and doors in a kitchen environment requires the accurate detection of rectangular objects in the depth image sequences.

From the stereo processing system, we obtain in each frame a disparity image $D \in \mathbb{R}^{640\times480}$, that contains for each pixel $(u,v)$ its perceived disparity $D(u,v) \in \mathbb{R}$. For more details on the camera system, in particular on the choice of a suitable texture projection pattern, we refer the interested reader to the recent work of Konolige (2010). The relationship between 2D pixels in the disparity image and 3D world points is defined by the projection matrices of the calibrated stereo camera, and can be calculated by a single matrix multiplication from the pixel coordinates and disparity.

We apply a RANSAC-based plane fitting algorithm for segmenting the dense depth image into planes. The next step is to find rectangles in the segmented planes. We start with a sampled candidate rectangle and optimize its pose and size iteratively, by minimizing an objective function that evaluates how accurately the rectangle candidate matches the points in the segmented plane. After the search converges, we determine the quality of the found rectangle by evaluating its pixel precision and recall. An example of the iterative pose fitting is given in the left image of Fig. 5: the rectangle candidate started in the lower left of the door, and iteratively converged to the correct pose and size of the door.

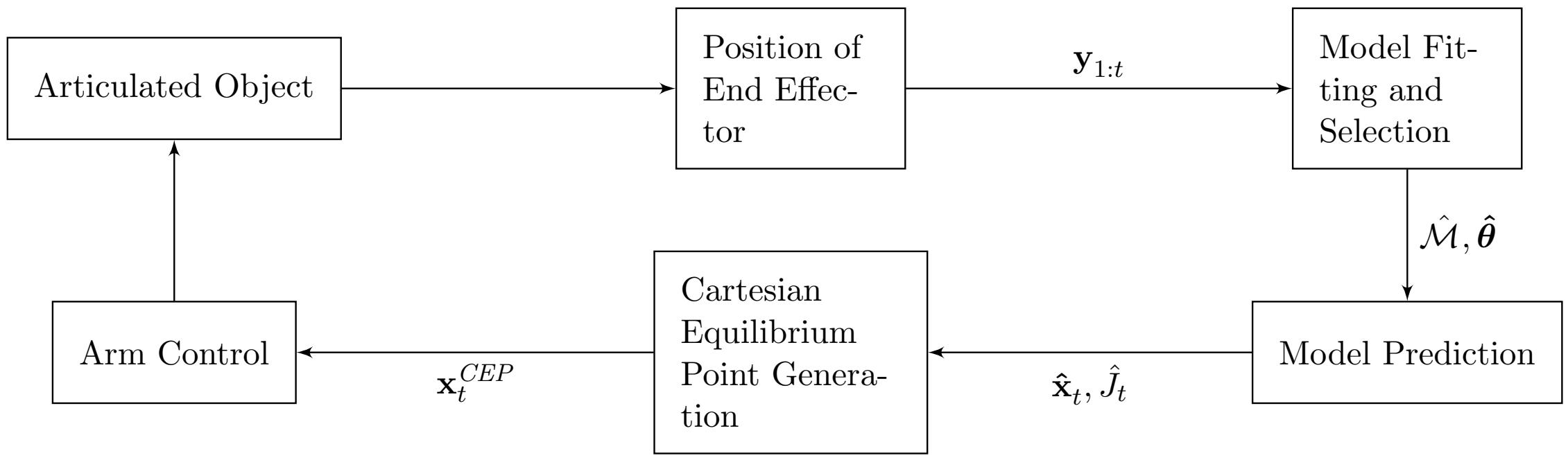

Figure 6: Overall control structure (Sturm et al., 2010). The robot uses the trajectory of its end effector to estimate a model of the articulated object. Subsequently, it uses this model for generating the next Cartesian equilibrium point.

Finally, for a sequence of depth images $D^{1:n}$, the detected rectangles need to be integrated into a set of consistent tracks, one for each visible rectangle. As a result, we obtain a set of pose sequences $\mathcal{D}_{\mathbf{y}} = (\mathbf{y}^1_{1:p}, \ldots, \mathbf{y}^n_{1:p})$ that we can use for model estimation and model selection. The middle and right image in Fig. 5 show the tracks that we obtained when observing a drawer and a door in a kitchen cabinet. More details on this approach have recently been described by Sturm et al. (2010).

### 4.3 Perception using the Joint Encoders of a Mobile Manipulation Robot

Next to visual observation of articulated objects, a mobile manipulation robot can also estimate the kinematic model while it physically interacts with the articulated object. By evaluating its joint encoders, the robot can compute the pose of its gripper by using the forward model of its manipulator. When the robot establishes firm contact with the handle of a cabinet door, the position of its end-effector directly corresponds to the position of the door handle. As a result, the robot can both sense the position of the handle as well control it by moving the manipulator.

The approach described in this section was developed in collaboration with Jain and Kemp from the Healthcare Robotics Lab at Georgia Tech. The robot that we use for this research is a statically stable mobile manipulator named Cody. It consists of two arms from MEKA Robotics and an omni-directional mobile base from Segway. As an end-effector, it uses a hook inspired by prosthetic hooks and human fingers, and is described in more detail in the recent work of Jain and Kemp (2009a). Furthermore, we used a PR2 robot from Willow Garage for additional experiments, using a standard 1-DOF gripper with two fingers, located in our lab.

Fig. 6 shows a block diagram of our approach. The robot observes the pose of its end effector in Cartesian space, denoted by $\mathbf{y} \in SE(3)$. While operating the mechanism, the robot records the trajectory $\mathbf{y}_{1:t}$ over time as a sequence of poses. From this partial trajectory, it continuously estimates the kinematic model of the articulated object, that the robot uses in turn to predict the continuation of the trajectory (Sturm et al., 2010).

To actively operate an articulated object with a robot, we use a trajectory controller that updates the Cartesian equilibrium point based on the estimated Jacobian of the kinematic model of the articulated object. This controller uses the kinematic model to generate Cartesian equilibrium point (CEP) trajectories in a fixed world frame, attached to the initial location of the handle. At each time step $t$, the controller computes a new equilibrium point $\mathbf{x}_t^{CEP}$ as

$$\mathbf{x}_t^{CEP} \quad = \quad \mathbf{x}_{t-1}^{CEP} + \mathbf{v}_t^{mechanism} + \mathbf{v}_t^{hook}, \tag{56}$$

where $\mathbf{v}_t^{mechanism}$ is a vector intended to operate the mechanism, and $\mathbf{v}_t^{hook}$ is a vector intended to keep the hook from slipping off the handle. The controller computes

$$\mathbf{v}_t^{mechanism} \quad = \quad L^{\,mechanism} \frac{\hat{J}_t}{\|\hat{J}_t\|} \tag{57}$$

as a vector of length $L^{\,mechanism} = 0.01\,\mathrm{m}$ along the Jacobian of the learned kinematic function of the mechanism, i.e.,

$$\hat{J}_t = \nabla f_{\hat{\mathcal{M}},\hat{\boldsymbol{\theta}}}(\mathbf{q})\big|_{\mathbf{q}=\hat{\mathbf{q}}^t}. \tag{58}$$

For $\mathbf{v}_t^{hook}$, we use a proportional controller that tries to maintain a force of 5 N between the hook and the handle in a direction perpendicular to $\hat{J}_t$. This controller uses the force measured by the wrist force-torque sensor of the robot. We refer the reader to the work of Jain and Kemp (2009b) for details about the implementation of equilibrium point control, and how it can be used to coordinate the motion of a mobile base and a compliant arm (Jain & Kemp, 2010).

The positional accuracy of the manipulator itself is very high, i.e., $\sigma_{\mathbf{y},\mathrm{pos}} \ll 0.01\,\mathrm{m}$. However, by using a hook as the end-effector, the robot cannot sense the orientation of the handle. As the manipulator is mounted on a mobile base, the robot can move around, and thus the positional accuracy of the sensed position of the hook in a global coordinate system (and thus including localization errors of the base) reduces to about $\sigma_{\mathbf{y},\mathrm{pos}} \approx 0.05\,\mathrm{m}$.

## 5. Experiments

In this section, we present the results of a thorough evaluation of all aspects of our approach. First, we show that our approach accurately and robustly estimates the kinematic models of typical household objects using markers. Second, we show that the same also holds for data acquired with the marker-less pose estimation using our active stereo camera system. Third, we show that our approach also works on data acquired with different mobile manipulation robots operating various pieces of furniture in domestic environments.

### 5.1 Microwave Oven, Office Cabinet, and Garage Door

For our first experiments, we use pose observations from three typical objects in domestic environments: the door of a microwave oven, the drawers of an office cabinet, and a garage door. The goal of these experiments is to show that our approach both robustly and accurately estimates link models, as well as the correct kinematic structure of the whole articulated object. In addition, we show that the range of the configuration space can be obtained during model estimation.

The motion of the microwave oven and the cabinet was tracked using a motion capture studio in collaboration with Pradeep and Konolige (Sturm et al., 2009), and the garage door using checkerboard markers. For each object, we recorded 200 data samples while manually articulating each object. For the evaluation, we carry out 10 runs. For each run, we sampled $n = 20$ observations that we use for fitting the model parameters. We used the remaining observations for measuring the prediction accuracy of the fitted model (10-folds cross-validation).

#### 5.1.1 Model Fitting

The quantitative results of model fitting and model selection are given in Table 2. As can be seen from this table, the revolute model is well suited for predicting the opening movement of the microwave door (error below 0.001 m) while the prismatic model predicts very accurately the motion of the drawer (error below 0.0016 m), which is the expected result. Note that also the revolute model can also explain the motion of the drawer with an accuracy of 0.0017 m, by estimating a rotary joint with a large radius. It should be noted that the flexible GP model provides roughly the same accuracy as the parametric models and is able to robustly predict the poses of both datasets (0.0020 m for the door, and 0.0017 m for the drawer). In the case of the simulated garage door, however, all parametric models fail whereas the GP model provides accurate estimates.

The reader might wonder now why the GP model alone does not suffice, as the GP model can represent many different types of kinematic models, including revolute and prismatic ones. However, even if the GP model fits all data, it is not the best choice in terms of the resulting posterior likelihoods. The GP model can – in some cases – be overly complex, and then over-fit the data at hand. This high complexity of the GP model is penalized by the BIC. In contrast, the specialized models have a smaller number of free parameters, and are therefore more robust against noise and outliers. Furthermore, they require less observations to converge. These experiments illustrate that our system takes advantage of the expert-designed parametric models when appropriate while keeping the flexibility to also learn accurate models for unforeseen mechanical constructions.

The learned kinematic models also provide the configuration range $C$ of the articulated object. For visualization purposes, we can now sample configurations from this range, and project them to object poses using the learned forward function. Fig. 7, Fig. 8, and Fig. 9 illustrate the learned configuration range for the door of the microwave oven, the garage door, and the two drawers of the office cabinet, respectively.

#### 5.1.2 Model and Structure Selection

After fitting the model candidates to the observed data, the next goal is to select the model that best explains the data, which corresponds to finding the model that maximizes the posterior probability (or minimizes the BIC score).

The right image in Fig. 7 shows the resulting graph for the microwave oven dataset, with the BIC score indicated at each edge. As expected, the revolute model is selected, because it has the lowest BIC score. Correspondingly, the right image in Fig. 8 shows the BIC scores for all edges for the garage door dataset, where the GP model gets selected.

| Dataset ↓ | | Rigid Model | Prismatic Model | Revolute Model | GP Model |
|---|---|---|---|---|---|
| Microwave | pos. error = | 0.3086 m | 0.1048 m | **0.0003 m** | 0.0020 m |
| ($\sigma_{\mathbf{z},\text{pos.}} = 0.002$ m, | orient. error = | 37.40° | 32.31° | **0.15°** | 0.16° |
| $\sigma_{\mathbf{z},\text{orient.}} = 2.0°$) | $\gamma$ = | 0.891 | 0.816 | **0.000** | 0.000 |
| Drawer | pos. error = | 0.0822 m | **0.0016 m** | 0.0018 m | 0.0017 m |
| ($\sigma_{\mathbf{z},\text{pos.}} = 0.002$ m, | orient. error = | 2.06° | **1.36°** | 1.60° | 1.09° |
| $\sigma_{\mathbf{z},\text{orient.}} = 2.0°$) | $\gamma$ = | 0.887 | **0.000** | 0.003 | 0.000 |
| Garage Door | pos. error = | 1.0887 m | 0.3856 m | 0.4713 m | **0.0450 m** |
| ($\sigma_{\mathbf{z},\text{pos.}} = 0.050$ m, | orient. error = | 14.92° | 10.79° | 10.34° | **0.93°** |
| $\sigma_{\mathbf{z},\text{orient.}} = 5.0°$) | $\gamma$ = | 0.719 | 0.238 | 0.418 | **0.021** |

Table 2: Prediction errors and estimated outlier ratios of the articulation models learned of a microwave oven, an office cabinet, and a real garage door.

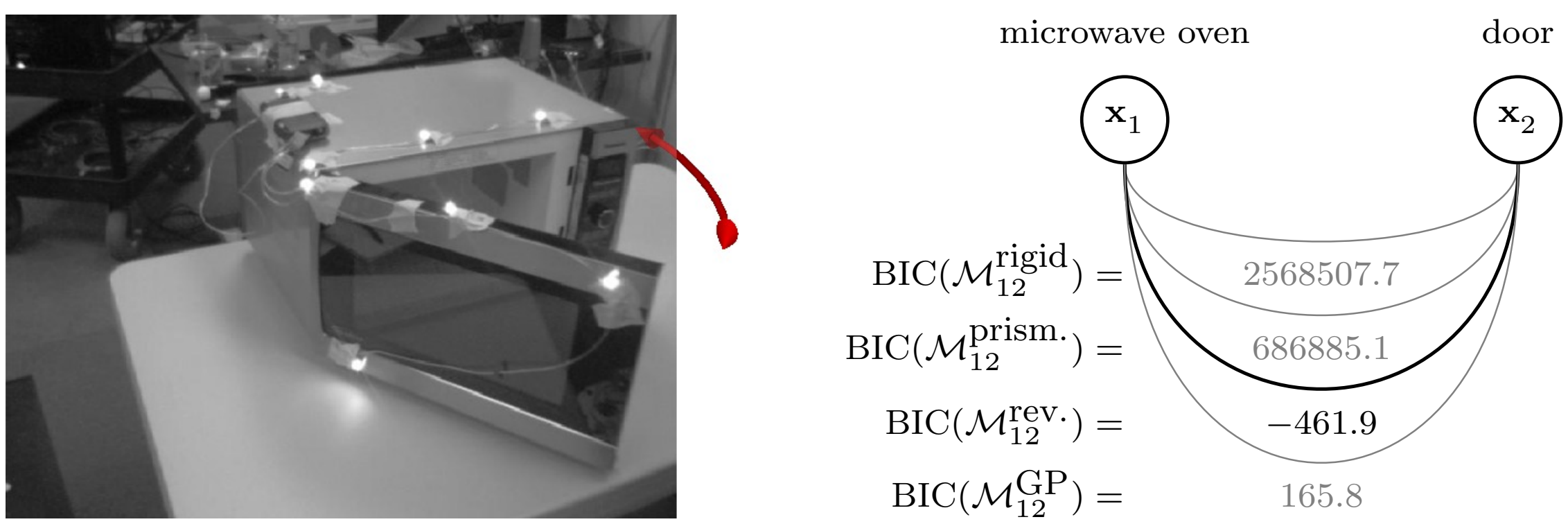

(a) (b)

Figure 7: Visualization of the kinematic model learned for the door of a microwave oven. (a) configuration range. (b) kinematic graph. The numbers on the edges indicate the BIC score of the corresponding model candidate.

A typical articulated object consisting of multiple parts is a cabinet with drawers as depicted in Fig. 9. In this experiment, we track the poses of the cabinet itself ($\mathbf{x}_1$), and its two drawers ($\mathbf{x}_2$ and $\mathbf{x}_3$). During the first 20 samples, we opened and closed only the lower drawer. Accordingly, a prismatic joint model $\mathcal{M}_{23}^{\text{prism.}}$ is selected (see top row of images in Fig. 9). When also the upper drawer gets opened and closed, the rigid model $\mathcal{M}_{12}^{\text{rigid}}$ is replaced by a prismatic model $\mathcal{M}_{12}^{\text{prism.}}$, and $\mathcal{M}_{23}^{\text{prism.}}$ is replaced by $\mathcal{M}_{13}^{\text{prism.}}$, resulting in the kinematic tree $E_G = \{(1,2),(1,3)\}$. Note that it is not required to articulate the drawers one after each other. This was done only for illustration purposes.

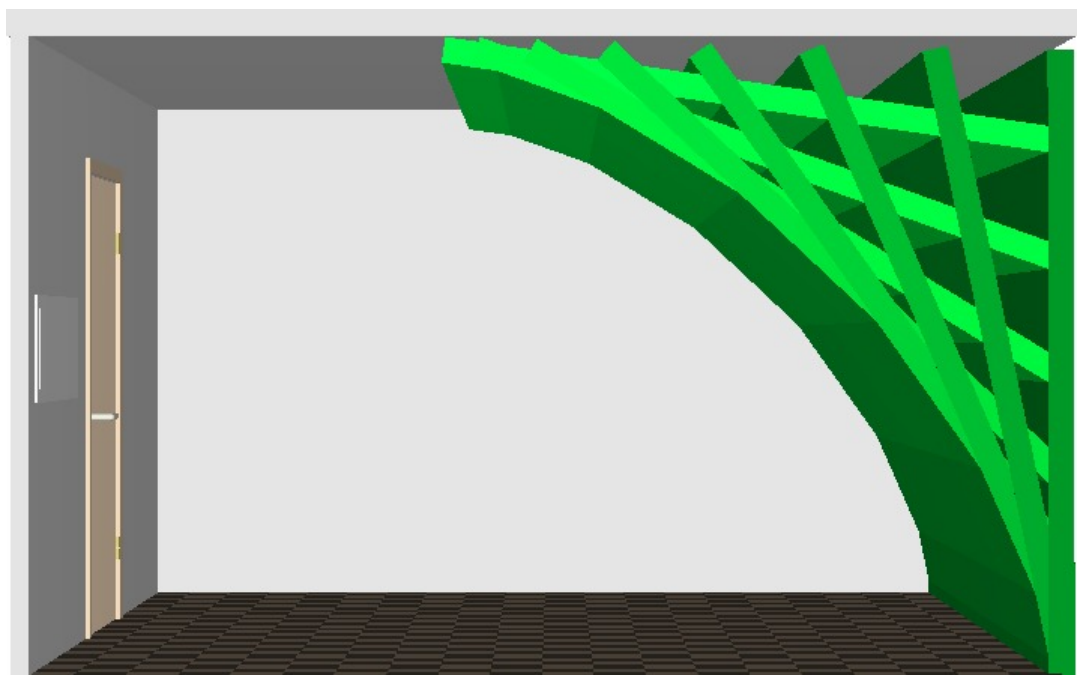

(a)

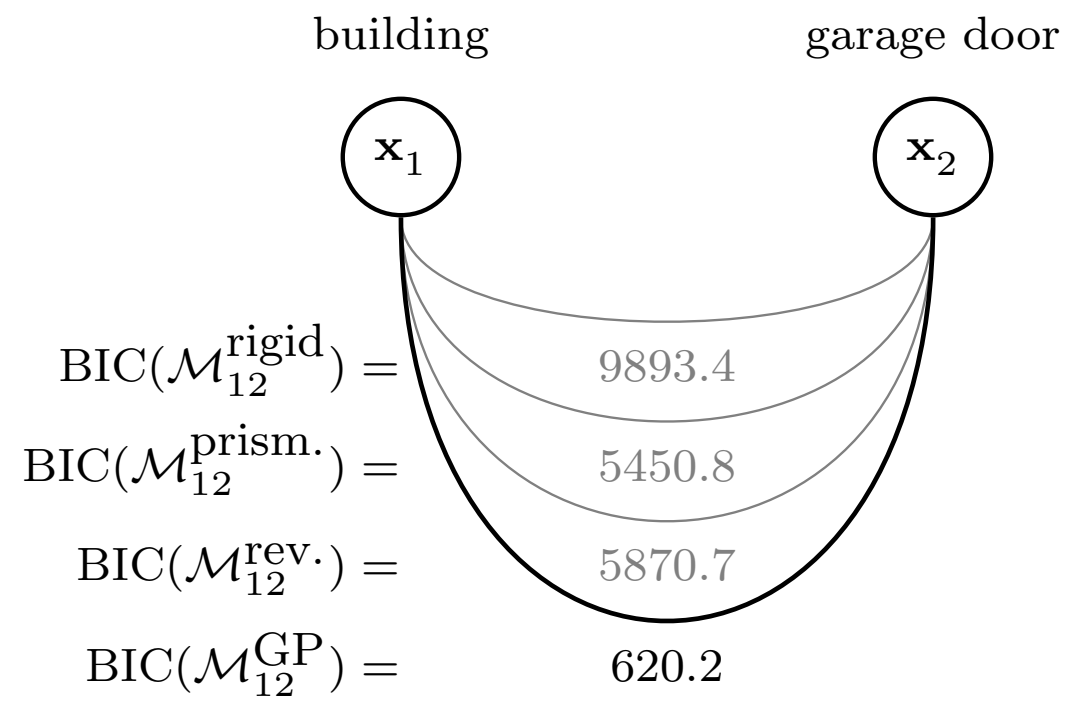

(b)

Figure 8: Visualization of the kinematic model learned for a garage door. (a) 10 uniformly sampled configurations. (b) kinematic graph.

#### 5.1.3 Multi-Dimensional Configuration Spaces

To illustrate that our approach is also able to find models with higher-dimensional configuration spaces with $d > 1$, we let the robot monitor a table that was moved on the floor. The robot is equipped with a monocular camera tracking an Artoolkit marker attached to the table. In this experiment, the table was only moved and was never turned, lifted, or tilted and therefore the configuration space of the table has two dimensions. Fig. 10 shows four snapshots during learning. Initially, the table is perfectly explained as a rigid object in the room (top left). Then, a prismatic joint model best explains the data since the table was moved in one direction only (top right). After moving sideways, the best model is a 1-DOF Gaussian process model that follows a simple curved trajectory (bottom left). Finally, the full planar movement is explained by a 2-DOF Gaussian process model (bottom right), that can model movements that lie on 2D surfaces.

#### 5.1.4 Additional Examples

We ran similar experiments on a large set of different articulated objects that typically occur in domestic environments, including office cabinets, office doors, desk lamps, windows, kitchen cabinets, fridges and dishwashers and a garage door. Four examples are given in Fig. 11. Videos of these (and all other) experiments are available on the homepage of the corresponding author[3]. These videos show both the original movie as well as an overlay of the inferred kinematic model. For these experiments, we attached checkerboards of different sizes to all movable parts, and used both a consumer-grade video camera and a low-cost laptop webcam for acquiring the image data. Our software also visualizes the learned articulation models in 3D and back-projects them onto the image to allow for easy visual inspection. The detected poses of the checkerboards are visualized as red/green/blue coordinate axes systems, and selected links between them are indicated using a colored connection. The software also displays the configuration range by generating poses in the

3. http://www.informatik.uni-freiburg.de/ sturm

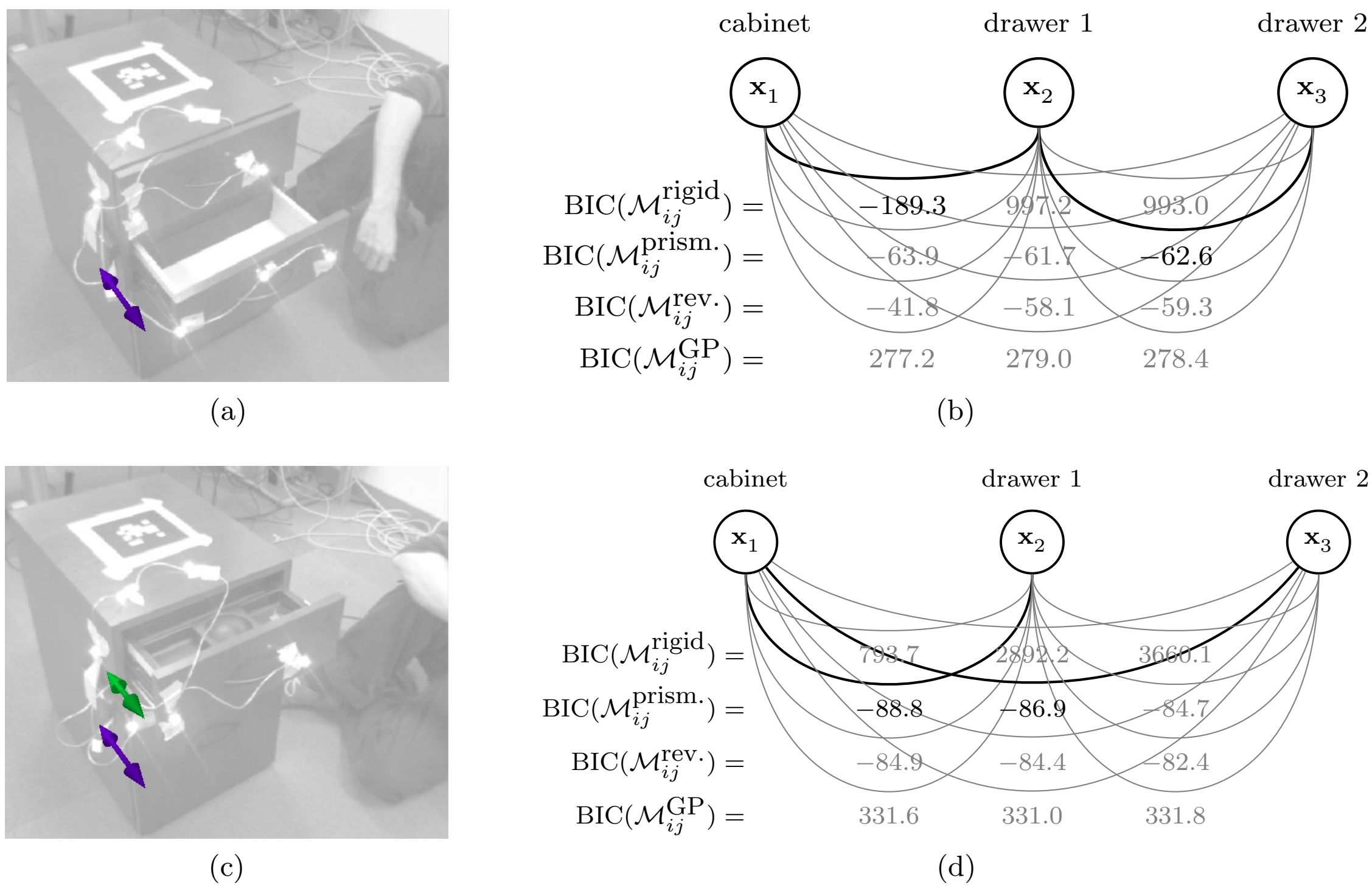

Figure 9: Incrementally estimating a model of two drawers of a cabinet. (a) Initially, only the lower drawer is opened and closed. (b) Corresponding kinematic graph. (c) Both drawers are opened and closed independently. (d) Corresponding kinematic graph.

estimated range. For revolute joints, it additionally indicates the rotation axis using a line and a surrounding circle.

From visual inspection of the objects in Fig. 11, one can see how accurate the model estimation works in conjunction with marker-based tracking: the motion of the drawers of the cabinet is well matched, and the rotation axes of the door hinge and the door handle are estimated very close to their true position. The upper part of the garage door moves in a slider in the ceiling, while the lower part is connected via a revolute joint. The resulting motion is clearly neither revolute nor prismatic, and consequently our approach selects the GP model. The desk lamp consists of two-bar links that keep the light housing always upright (or, loosely speaking, rigid in orientation), but move in the positional domain along a circle. This link type can be well explained by the GP model. The existence of these objects shows the necessity to supply a domestic service robot with such a general, non-parametric model that can deal with a wide variety of different articulated objects. It should be noted that the majority of articulated objects in domestic environments will consist of revolute and prismatic joints which can be more robustly estimated using parametrized models. This motivates our approach that enables a robot to fit both parametric and nonparametric models at the same time and compare them in terms of posterior likelihoods in a consistent model selection framework.

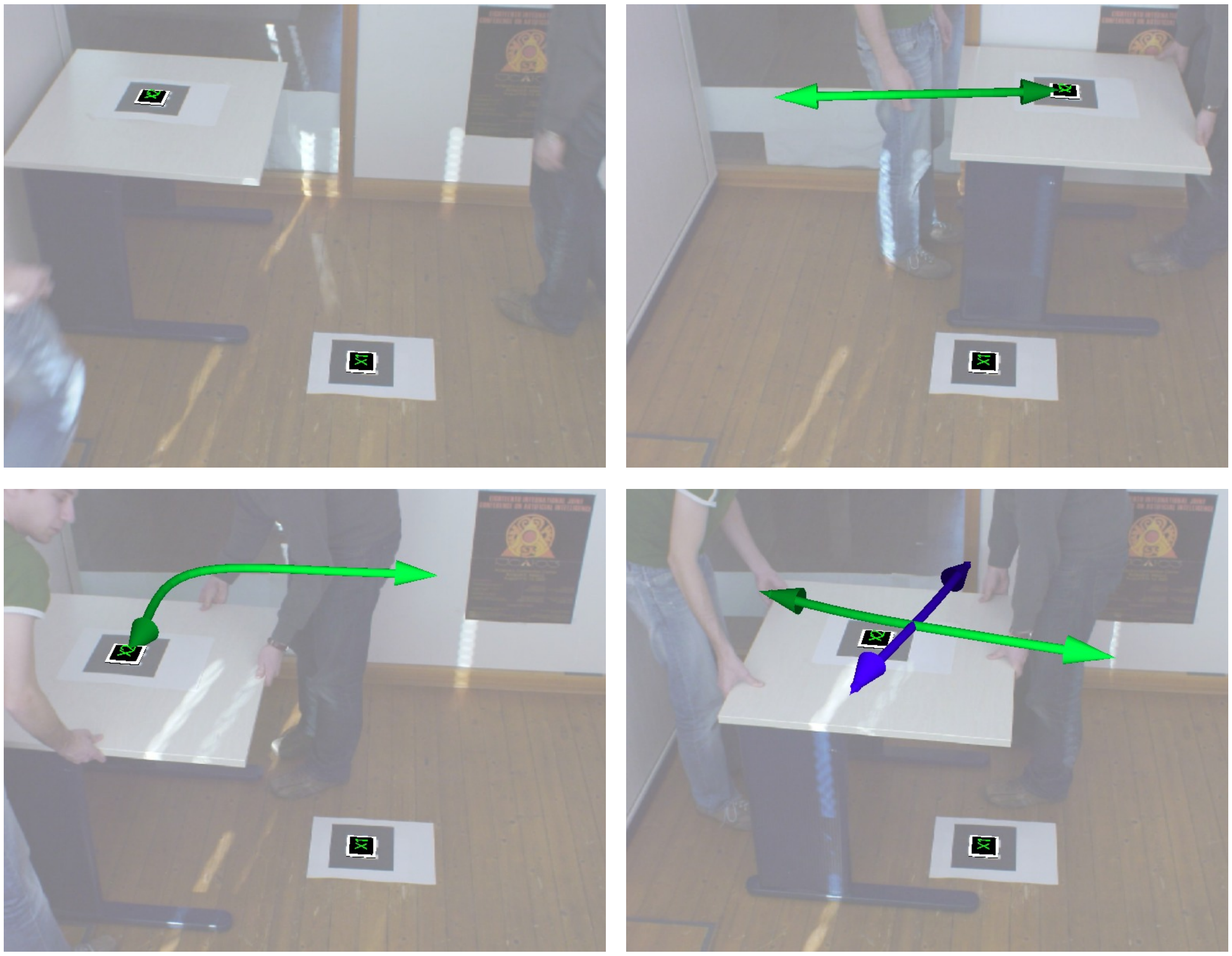

Figure 10: Learning a model for a table moving on the ground plane. The arrows indicate the recovered manifold of the configuration space.

Another interesting object is a car, as its doors and windows have both tree- and chain-like elements. In Fig. 12, we observed the motion of the driver's door and window. After the first few observations, our approach estimates the structure to be rigid, and links both the door and the window in parallel to the car body. After we open the window to the half, our approach attaches the driver's window to the door, and selects a prismatic model. Surprisingly to us, when we open the window further (and thus acquire more observations), our approach switches to a revolute model for the driver's window associated with a large radius ($r = 1.9\,\mathrm{m}$). By looking carefully at the data and the car, we can confirm that the window indeed moves on a circular path, which is due to its curved window glass. Finally, after the driver closes the door, also a revolute model for the link between the car body and the door is selected.

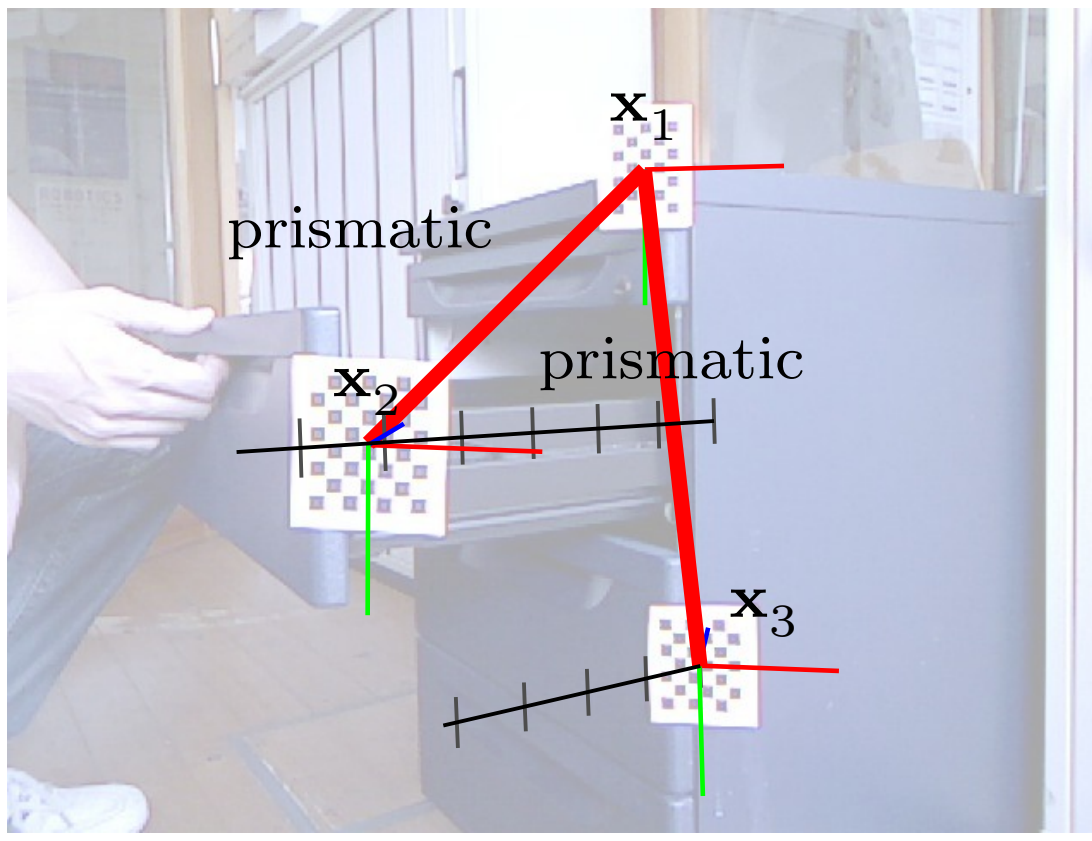

(a) Office cabinet

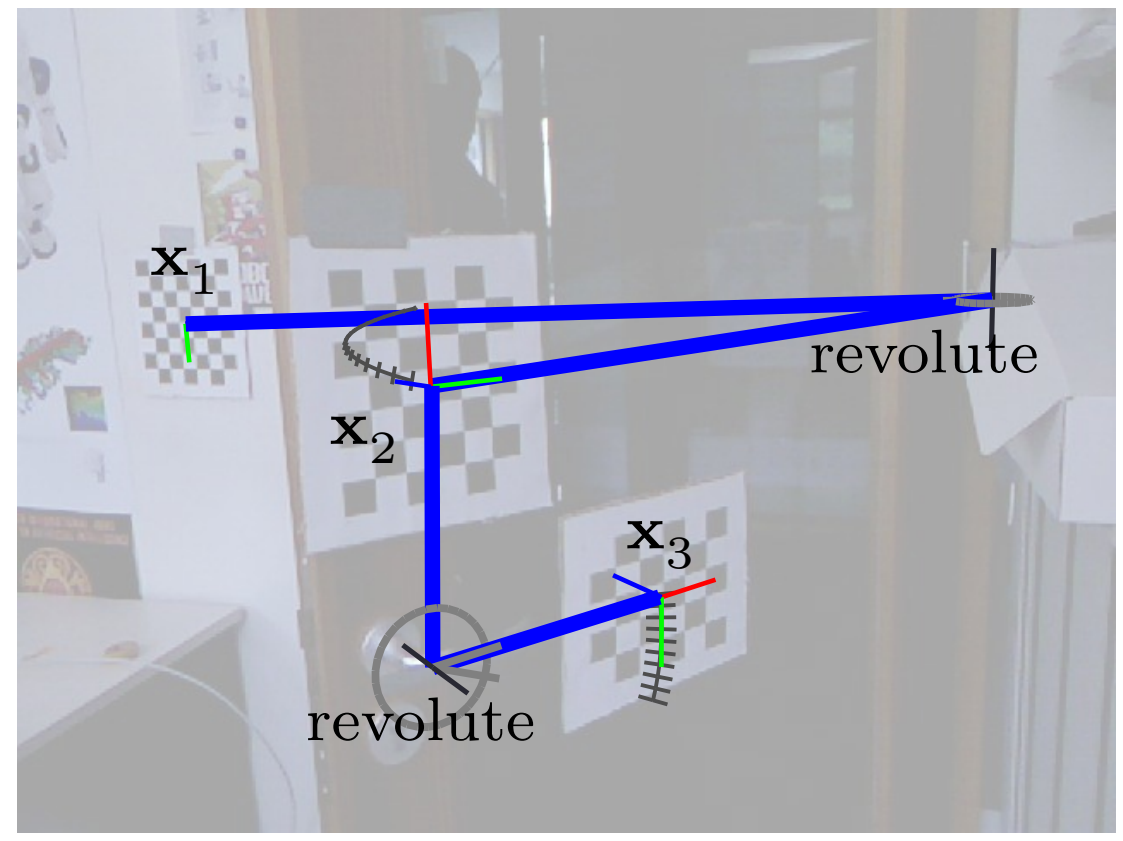

(b) Room door

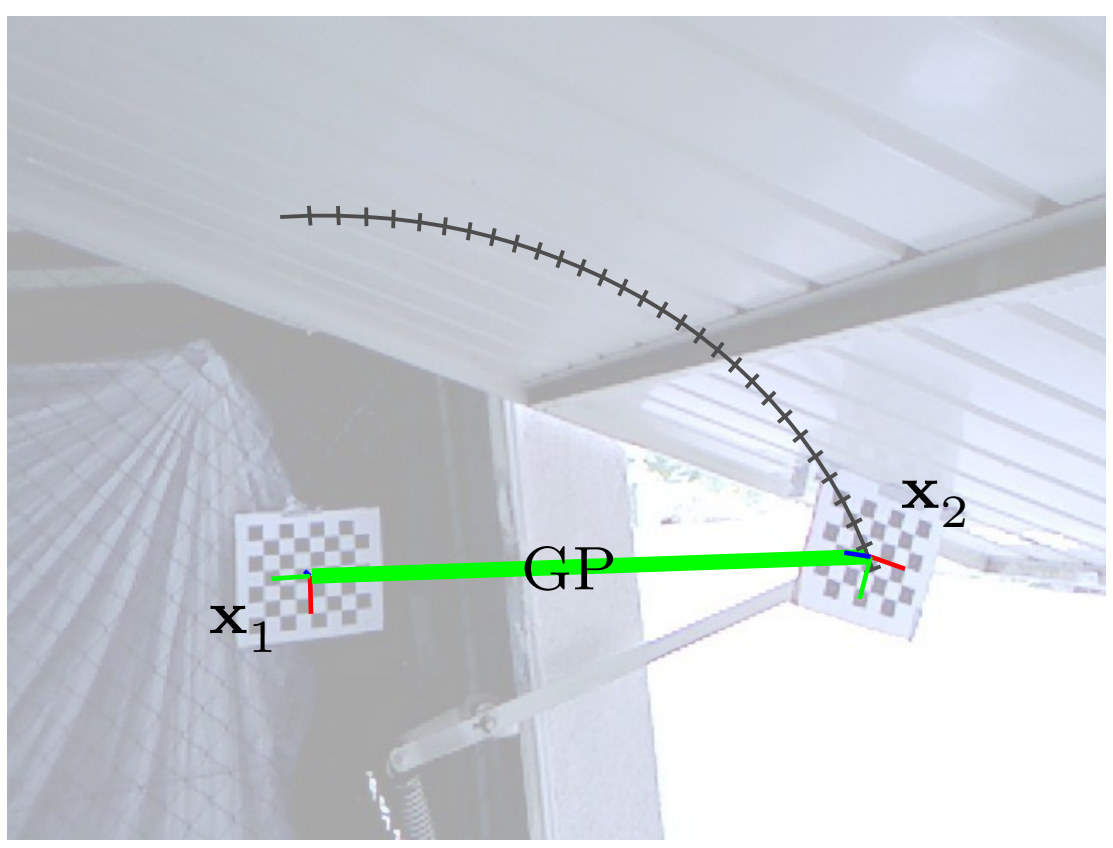

(c) Garage door

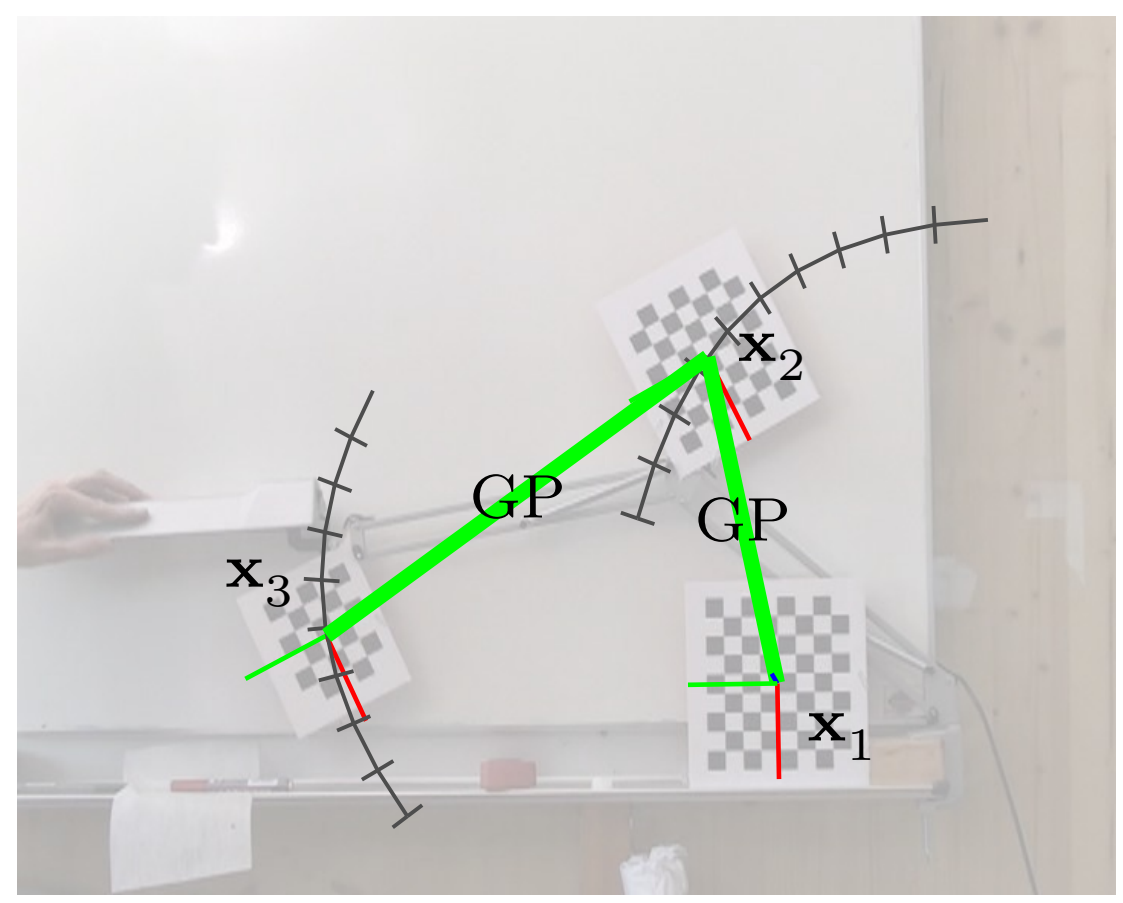

(d) Desk lamp

Figure 11: Visualization of the learned articulation models for several further domestic objects. (a) cabinet with two drawers, (b) room door including handle, (c) garage door, (d) desk lamp with two-bar links.

We conclude from these results, that our approach is able to estimate the kinematic parameters and the kinematic structure of different household objects at high accuracy, i.e., the prediction error of the learned models is around 0.001 m and 1° for objects tracked in a motion capture studio, and around 0.003 m and 3° for checkerboard markers. At this accuracy, the learned models are well suited for mobile manipulation tasks.

### 5.2 Evaluation of Marker-Less Model Estimation

The goal of our next set of experiments is to show that the kinematic models can be learned in certain environments without requiring artificial markers. In particular, we focus here on kitchen environments with rectangular cabinet fronts, and employ our pose detector described previously (Sturm et al., 2010).

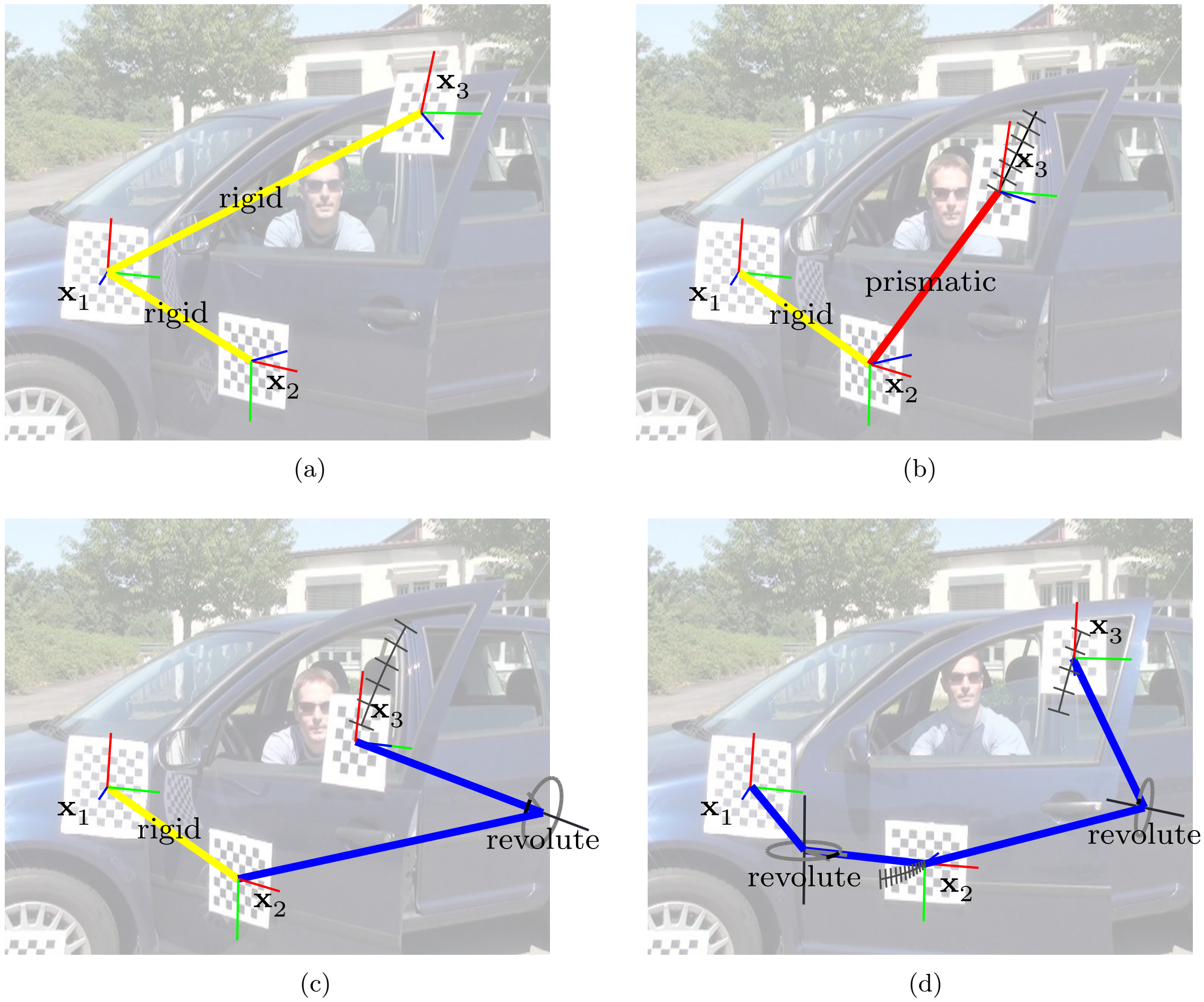

Figure 12: Snapshots of the learning process when incrementally observing the motion of a car door and its window from camera images. Due to the shape of the glass the driver's window actually moves on a circular arc with radius $r = 1.9$ m. Images taken after (a) 10, (b) 40, (c) 60, and (d) 140 pose observations.

In a first experiment carried out in a motion capture studio, we evaluated our detector and found that it detected the cabinet drawer in more than 75% of the images up to a distance of 2.3 m from the camera.

We evaluated the robustness of our articulation model learner on detailed logfiles of both a door (0.395 m× 0.58 m) and a drawer (0.395 m× 0.125 m) of a typical kitchen interior. We repeatedly opened and closed these objects in approximately 1 m distance of the robot; in total, we recorded 1,023 and 5,202 images. We downsampled these logs stochastically to 100 images, and ran pose estimation, model estimation, and structure selection for 50 times. The outcome of the model selection process, and the accuracy of the selected model is depicted in Fig. 14 for the door dataset.

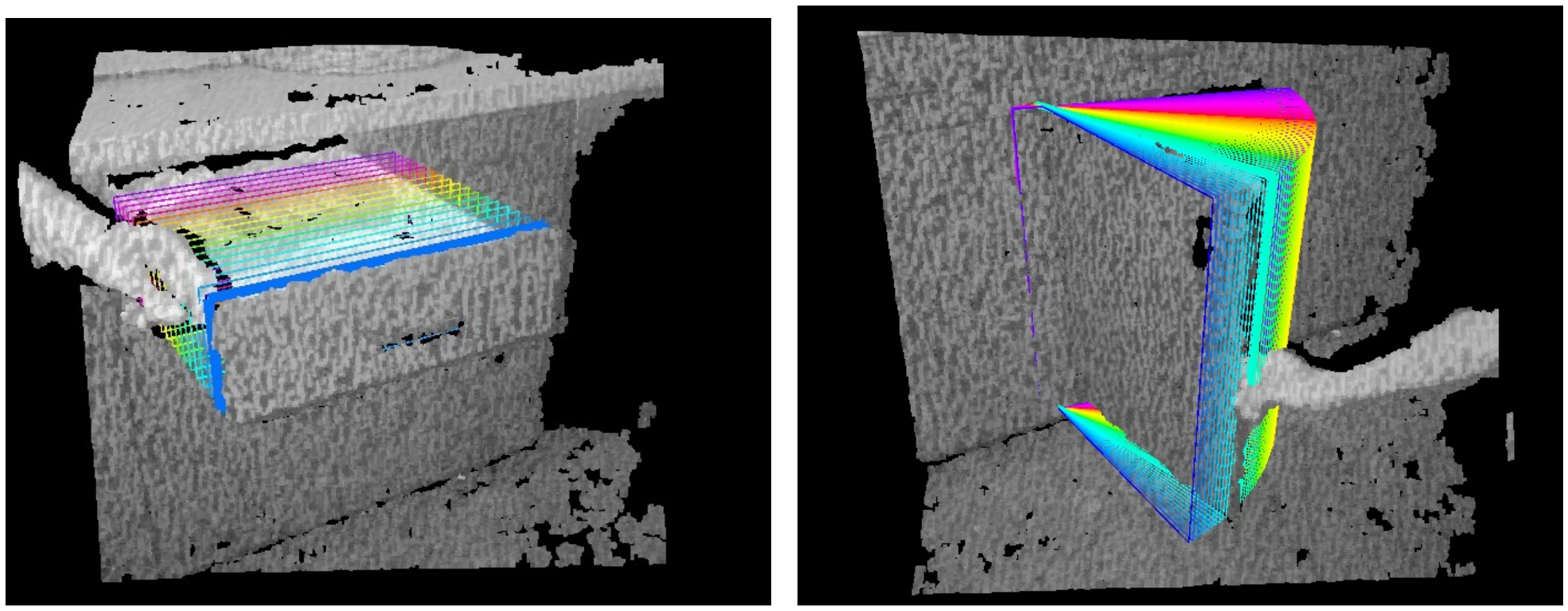

(a) (b)

Figure 13: Articulation model learned from observing a drawer (a) and a door (b).

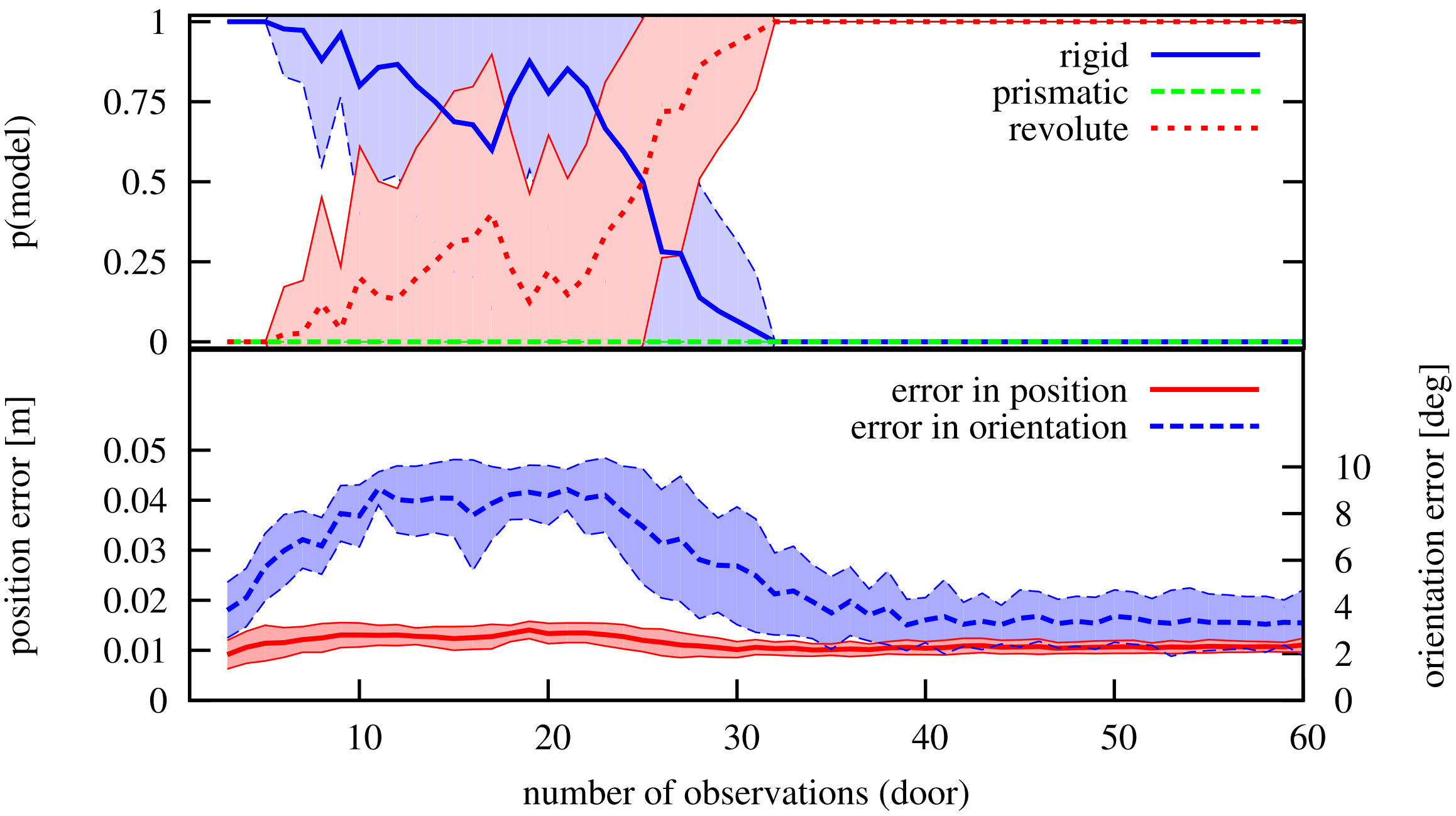

Figure 14: Evaluation of the articulation models learned for a cabinet door, averaged over 50 runs. The plot at the top shows the probability of the articulation model templates, the plot at the bottom shows the prediction error and standard deviation of the learned model.

For both datasets, we found that roughly for the first 10 observations, mostly the rigid model is selected, as no substantial motion of the drawer or door was yet detected. The more observations are added to the track, the higher the error between the (rigid) model predictions and the observations becomes. As a result, the prismatic and revolute models are selected more frequently. After 30 observations, model selection has converged in all cases to the true model.

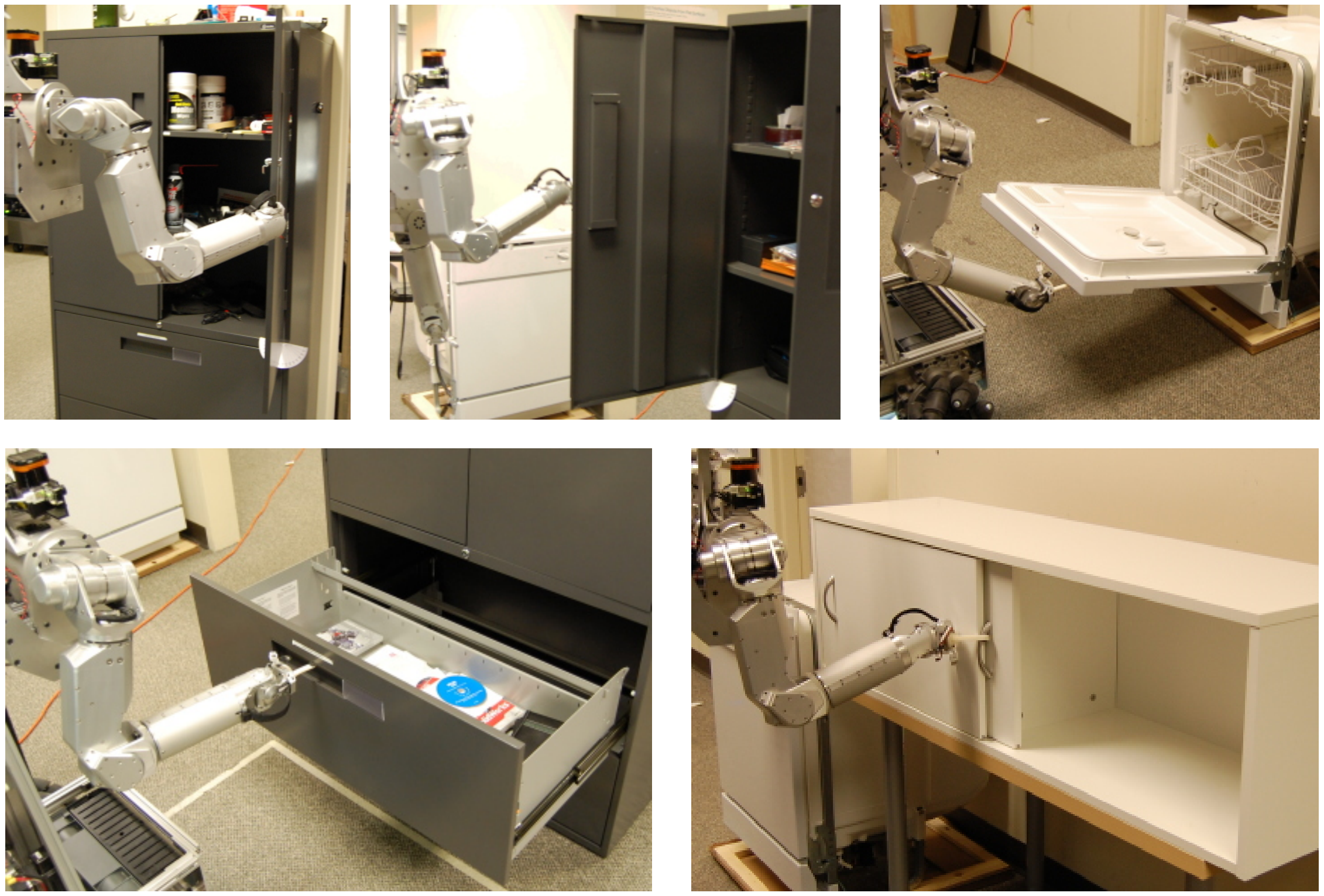

Figure 15: Images showing the robot Cody at Georgia Tech operating the five mechanisms using the approach described in Section 4.3. The objects are (from left to right): a cabinet door that opens to the right, a cabinet door that opens to the left, a dishwasher, a drawer, and a sliding cabinet door. Images courtesy of Jain and Kemp.

The models learned of the drawer datasets have a predictive accuracy of approximately 0.01 m and 7°; and 0.01 m and 3.5° for the door dataset. Although the predictive accuracy of the learned models is slightly lower in comparison with the marker-based tracking systems due to the higher noise of the tracking system, the learned models are at this accuracy usable for a mobile manipulator operating these objects.

### 5.3 Operating Articulated Objects with Mobile Manipulators

In this section, we show that real robots can utilize our approach to learn the kinematic models of objects for active manipulation. Here, control of the arm was done using the equilibrium point control as described in Section 4.3, which was the result of a collaboration with Jain and Kemp (Sturm et al., 2010). The experiments were conducted on two different platforms, the robot "Cody" and a PR2 robot.

### 5.3.1 Task Performance

We evaluated the performance of our approach on five different mechanisms using the robot Cody: a cabinet door that opens to the right, a cabinet door that opens to the left, a dishwasher, a drawer, and a sliding cabinet door. We performed eight trials for each mechanism. The robot started approximately 1 m from the location of the handle. We manually specified the grasp location by selecting a point in a 3D point cloud recorded by the robot, an orientation for the hook end effector, and the initial pulling direction. The task for the robot was to navigate up to the mechanism and operate it, while learning the articulation model using the methods described in Section 3.1. We deemed a trial to be successful if the robot navigated to the mechanism and opened it through an angle greater than $60°$ for revolute mechanisms or 0.3 m for prismatic mechanisms.

Fig. 15 shows the robot after it has pulled open each of the five mechanisms in one of the respective trials. The robot successfully opened the 3 rotary mechanisms in 21 out of 24 trials and the 2 linear mechanisms in all 16 trials. The robot was able to open the doors more than $70°$, and to estimate their radii on average with an error below 0.02 m. Further, the robot pulled open the drawer and the sliding cabinet repeatedly on average over 0.49 m. Overall the robot was successful in 37 out of 40 trials (92.5%).

All three failures were due to the robot failing to hook onto the handle prior to operating the mechanism, most likely due to odometry errors and errors in the provided location of the handle. In our experiments, we did not observe that the model learning caused any errors. In principle, however, the hook could slip off the handle if a wrong model had been estimated.

### 5.3.2 Model Fitting and Selection from End-Effector Trajectories

Fig. 1 and Fig. 16 show examples of the PR2 robot operating several articulated objects common to domestic environments, i.e., a fridge, a drawer, a dishwasher door, the tray of a dishwasher, and the valve of a heater. For these experiments, we did not use feedback control as described in Section 4.3 but tele-operated the manipulator manually. First, we recorded a set of trajectories by guiding the manipulator to operate various articulated objects. During execution, we played these trajectories back using a different implementation of equilibrium point control available on the PR2 platform, and recorded the end-effector trajectories of the robot. We used these trajectories subsequently to learn the kinematic models. Finally, we visualized these models by superimposing them on images taken by a calibrated wide-angle camera mounted on the head of the robot, see Fig. 16. In our experiments, our approach always selected the correct model candidate. One can easily verify by visual inspection that our approach estimates the kinematic properties (like the rotation axis or the prismatic axis) very accurately.

These experiments show that robots can successfully learn accurate kinematic models of articulated objects from end-effector trajectories by using our approach. With the PR2, we achieved an average predictive accuracy of the learned models below 0.002 m (in terms of residual error of the observed trajectory with respect to the learned model), which is more than sufficient for using our models for mobile manipulation tasks in domestic settings.

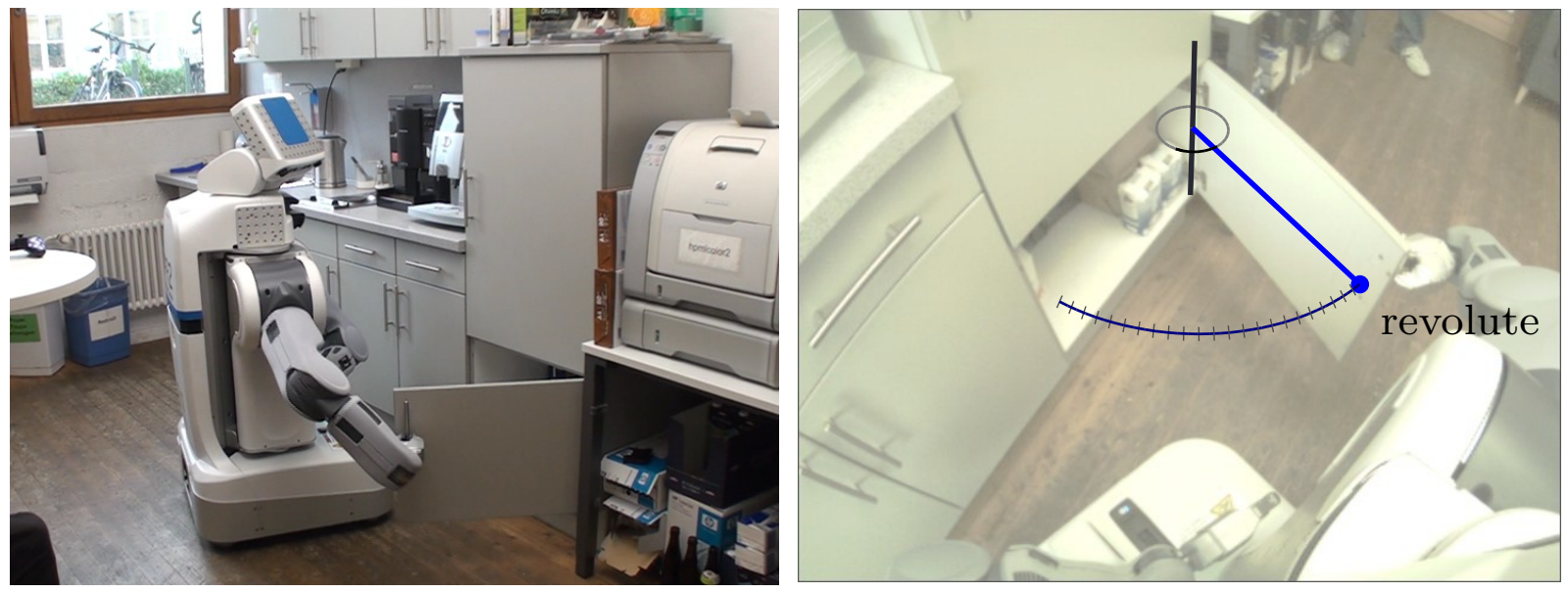

(a) Cabinet door

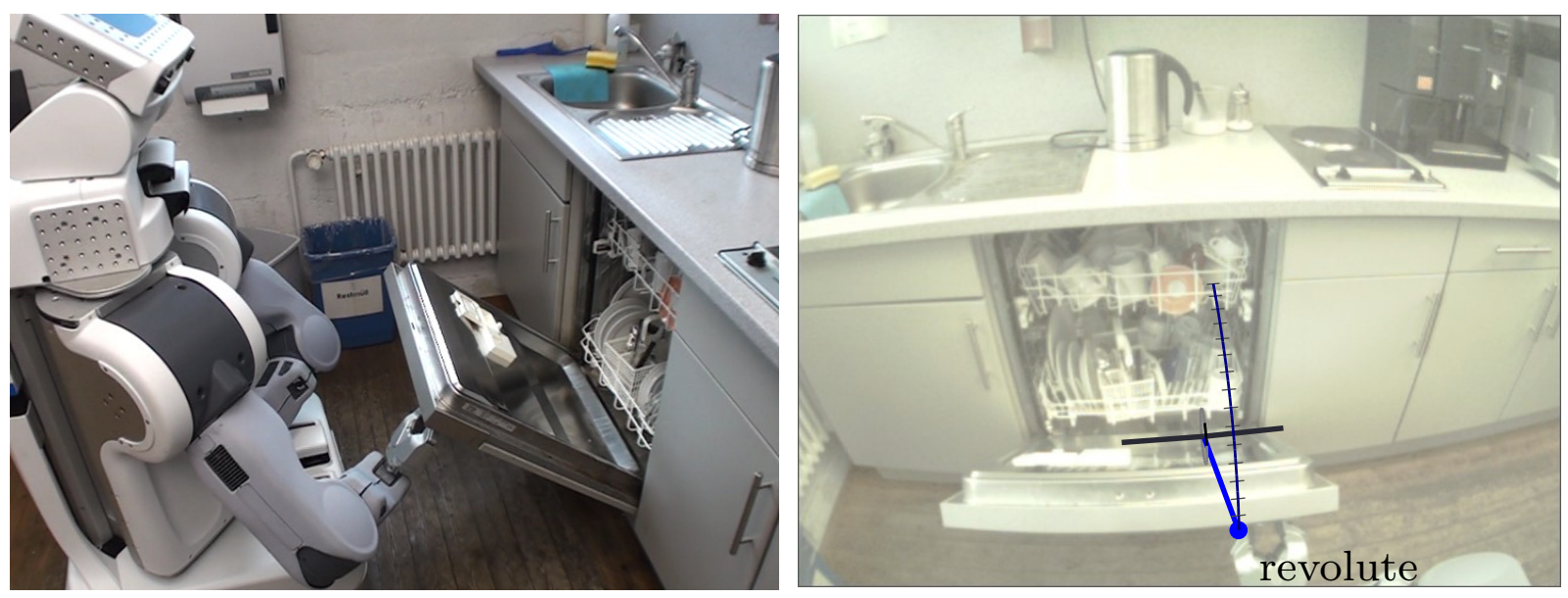

(b) Dishwasher door

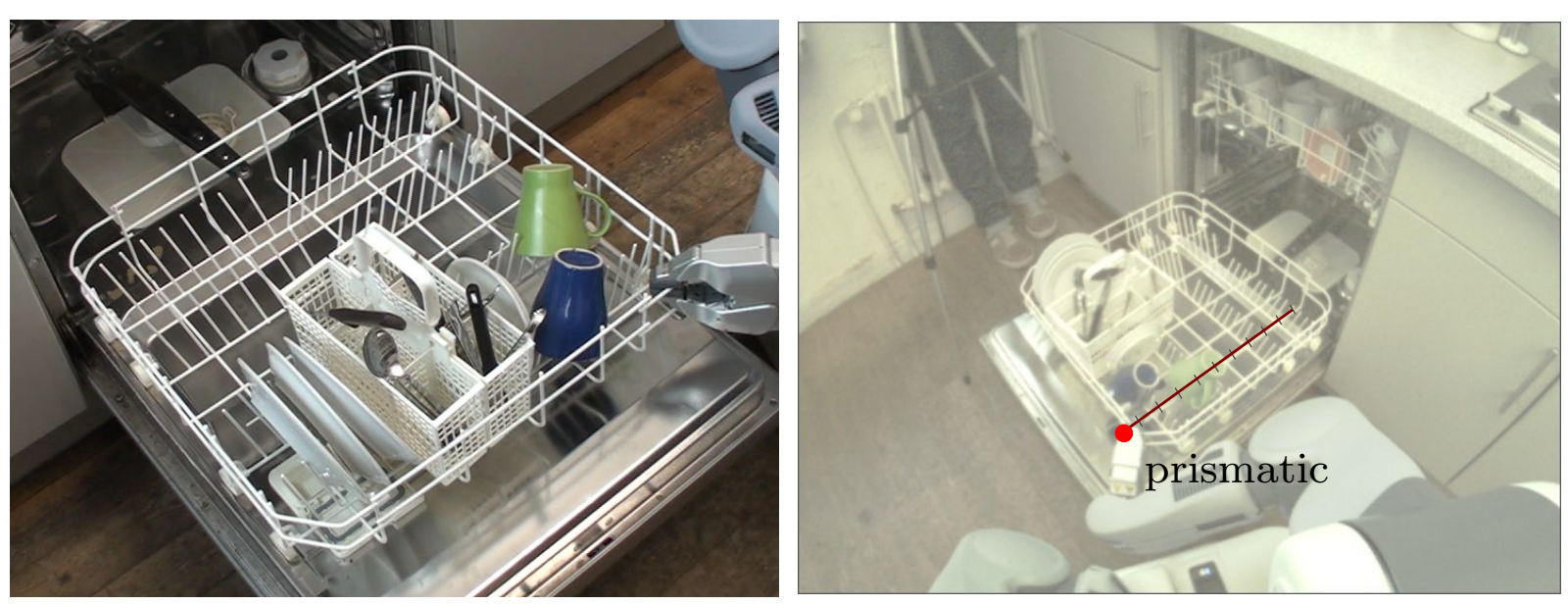

(c) Dishwasher tray

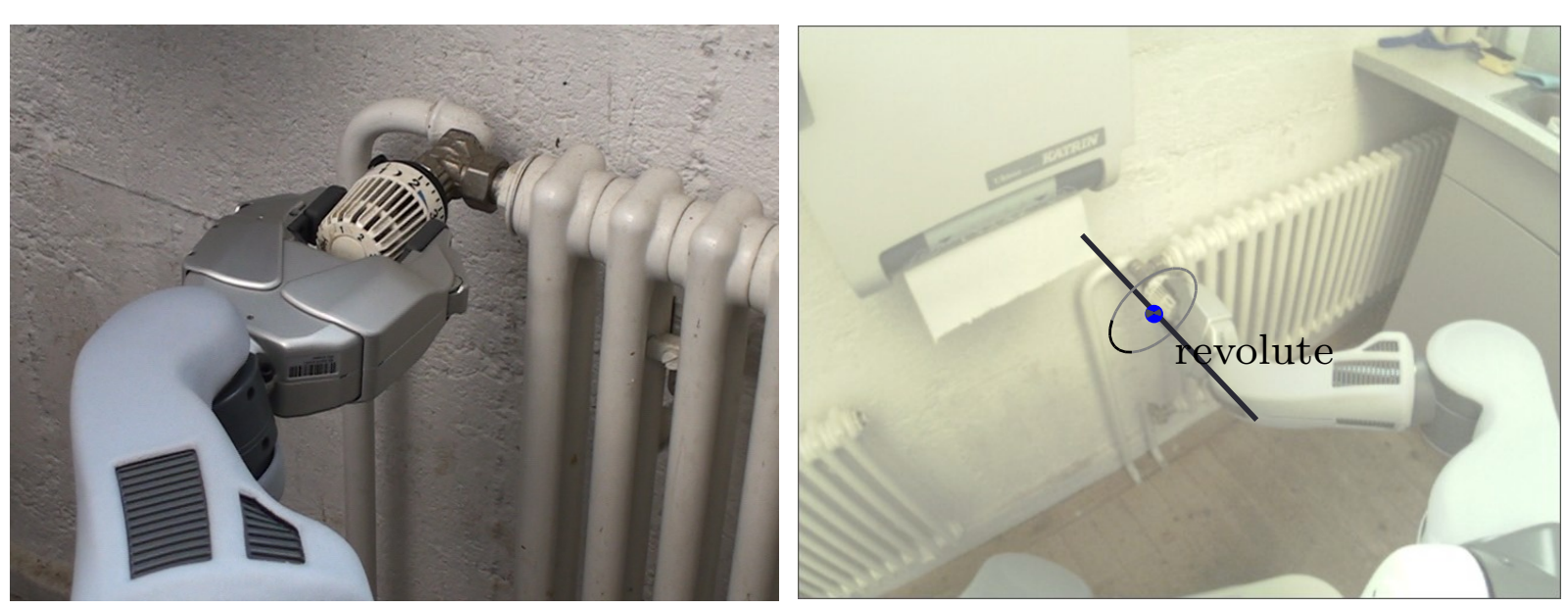

(d) Valve of a heater

Figure 16: A PR2 robot learns the kinematic models of different pieces of furniture by actuating them using its manipulator. Objects from top to bottom: fridge, cabinet door, drawer, dishwasher door, dishwasher tray, water tap, valve of a heater.

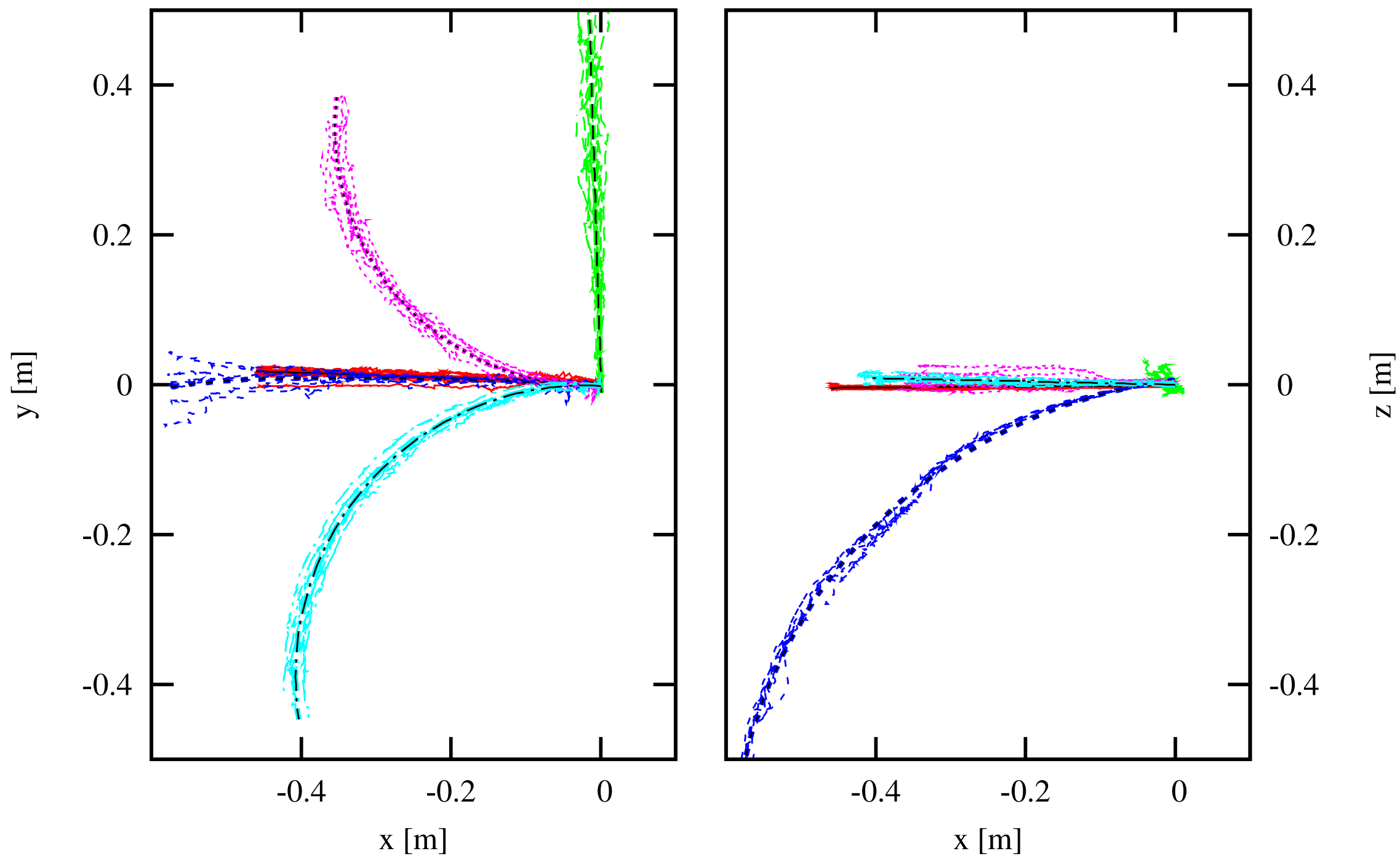

Figure 17: These plots show the observed trajectories and the 5 recovered models when minimizing the overall BIC using our approach. Trajectories assigned to the same model are depicted in the same color.

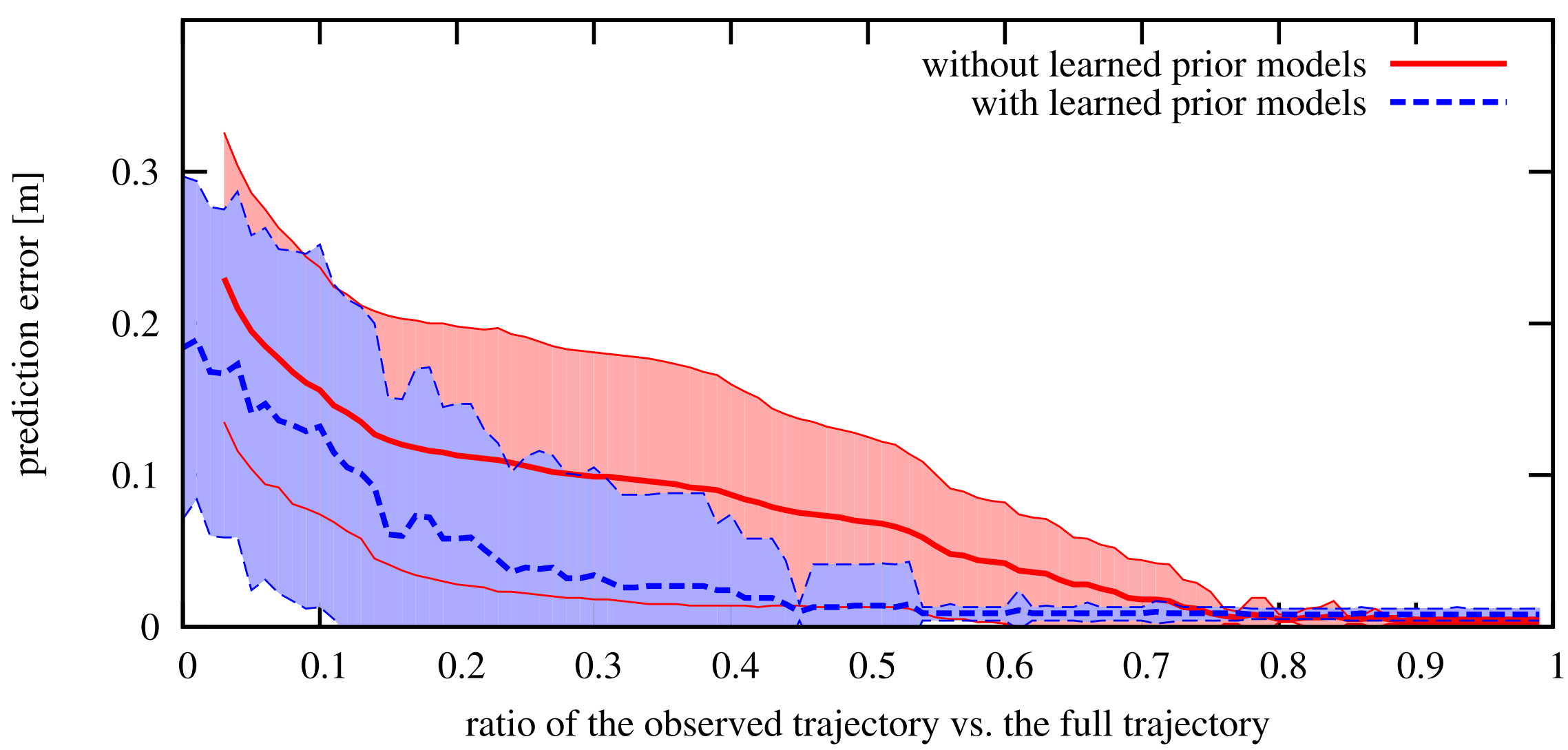

Figure 18: This graph shows the average prediction error (line) and standard deviation (shaded area) of the learned model on the full trajectory with and without prior information.

### 5.4 Improving Model Estimation Based on Experience

In the experiments described in the previous section, we learned the kinematic models for the kitchen furniture independent of each other. By using the approach described in Section 3.1 on data from Cody, we exploit the correlation between the models of different objects by searching for the set of model clusters that maximize the posterior probability. Fig. 17 shows the result of this experiment. The colors indicate the cluster to which the trajectories have been assigned to. Our approach correctly recognized that the robot had operated 5 different mechanisms and assigned the 37 different trajectories correctly to the corresponding models.

We measured the average prediction error with and without learning prior models (see Fig. 18), using leave-one-out cross-validation and a randomized ordering of the trajectories. We found that the prior models reduce the prediction error considerably, especially if the new trajectory is only partially observed. When 30% to 70% of the new trajectory have been observed, the prediction error is reduced by a factor of three and more. As a result, the robot comes up with a substantially more accurate model early and can utilize this knowledge to better control its manipulator.

Throughout all experiments on Cody, we used a fixed noise term of $\sigma_{\mathbf{z},\mathrm{pos}} = 0.05\mathrm{m}$. This accounts for inaccuracies in the observation of the end effector position, due to variations in the hooking position, and small errors in the kinematic forward model and robot base localization. We found in repeated experiments that in the range between $0.02\mathrm{m} \leq \sigma_{\mathbf{z},\mathrm{pos}} \leq 0.20\mathrm{m}$, the results are similar to our previous results obtained with $\sigma_{\mathbf{z},\mathrm{pos}} = 0.05\mathrm{m}$. Only for significantly smaller values of $\sigma_{\mathbf{z},\mathrm{pos}}$ more models are created, for example due to small variations of the grasping point and other inaccuracies. For much larger values, observations from different mechanisms are clustered into a joint model. Thus, our results are insensitive to moderate variations in the observation noise $\sigma_{\mathbf{z},\mathrm{pos}}$.

This experiment illustrates that our approach enables a mobile robot to learn from experience or exploit prior information when manipulating new objects. The experience increases the prediction accuracy by a factor of approximately three.

### 5.5 Detecting Kinematic Loops

In our final set of experiments, we evaluated our approach on objects containing kinematic loops. The goal of these experiments is to show that our approach can estimate correctly both the kinematic connectivity, as well as the correct number of DOFs.

For that purpose, we used the first four segments of a yardstick. This results in an open kinematic chain consisting of three revolute joints (see top left image of Fig. 19). This object has three DOFs, as all revolute joints are independent of each other. In a second experiment, we taped the fifth segment of the yardstick together with the first one. This creates an kinematic loop, see top right image of Fig. 19: the resulting object consists of four revolute joints each having a single DOF. The resulting mechanism has effectively only a single DOF. We articulated the objects manually, and recorded object pose datasets with $|\mathcal{D}_{\mathbf{y}}| = 200$ samples each using checkerboard markers.

The second and the third row of Fig. 19 visualize the learned kinematic model for the open and the closed kinematic model, respectively, while the fourth row shows the kinematic structure of the learned model. From this figure, it can be seen that our ap-proach correctly recognizes that the open kinematic chain consists of three revolute links ($\mathcal{M}_{12}^{\text{rev.}}, \mathcal{M}_{23}^{\text{rev.}}, \mathcal{M}_{34}^{\text{rev.}}$), having three DOFs $\mathbf{q} = (q_1, q_2, q_3)$ in total. For the closed kinematic chain, our approach selects four revolute links ($\mathcal{M}_{12}^{\text{rev.}}, \mathcal{M}_{23}^{\text{rev.}}, \mathcal{M}_{34}^{\text{rev.}}, \mathcal{M}_{14}^{\text{rev.}}$), and correctly infers that the object only exhibits a single DOF $\mathbf{q} = (q_1)$.

We also analyzed the progression of model selection while the training data is incorporated. The left plot of Fig. 20 shows the DOFs of the learned kinematic model for the open kinematic chain. Note that we opened the yardstick segment by segment, therefore the number of DOFs increases step-wise from zero to three. The right plot shows the estimated number of DOFs for the closed kinematic chain: our approach correctly estimates the number of DOFs to one already after the first few observations.

In more detail, we have analyzed the evolution of the BIC scores and the runtime of the different approaches for the closed kinematic chain in Fig. 21. The plot in the top shows the evolution of the BIC scores of all possible kinematic structures. We have colorized the curves corresponding to the spanning tree solution (solid red), heuristic search (dashed blue) and the global optimum (dotted green). The spanning tree solution that we use as the starting point for our heuristic search is on average 35.2% worse in terms of BIC than the optimal solution. In contrast, the BIC of the heuristic search is only 4.3% worse, and equals the optimal solution in 57.5% of the cases. The time complexity of computing the spanning tree is independent of the number of training samples, see bottom plot in Fig. 21. In contrast to that, the evaluation of kinematic graphs requires for each kinematic structure under consideration the evaluation of whole object poses, and thus is linear in the number of training samples $n$. The heuristic search only evaluates kinematic graphs along a trace through the structure space. As a result, for the yardstick object with $p = 4$ object parts, the heuristic search requires on average 82.6% less time than the full evaluation.

We conducted similar experiments on other objects containing kinematic loops and reduced DOFs. Two examples are depicted in the first row of Fig. 22: an artificial object consisting of four parts but only a single revolute joint, and a common domestic step ladder consisting of two revolute joints with only a single, shared DOF. In all cases, our approach was able to correctly estimate both the kinematic parameters and the correct number of DOFs.

With these experiments, we have shown that our approach is able to detect closed chains in articulated objects, and correctly estimates the correct number of DOFs. As loop closures (or reduced DOFs) reduce the configuration space of an object significantly, this is valuable information for a mobile manipulator, for example while reasoning about possible configurations of an object.

#### 5.5.1 Evaluation of Model Selection Robustness

Finally, we investigated the influence of the choice of the observation noise variable $\Sigma_{\mathbf{z}}$ on the model selection process on artificial data. For this analysis, we sampled noisy observations from a revolute model with a true observation noise of $\sigma_{\mathbf{z},\text{pos}}^{\text{true}} = 0.05$. On the resulting observation sequence, we fitted the candidate models, and selected the best model. We repeated this experiment for 10 independent runs and evaluated the mean and the standard deviation. When the number of training samples $n$ is kept fixed, then a higher noise assumption favors the selection of simpler models, and vice versa. Fig. 23 illustrates this dependency: for $n = 50$ and an assumed noise level of $\sigma_{\mathbf{z},\text{pos}} \leq 0.02$, the GP model is selected. Between $0.02 \leq \sigma_{\mathbf{z},\text{pos}} \leq 0.2$, the revolute model yields the best trade-off between model complexity and data likelihood. Above $0.2 \leq \sigma_{\mathbf{z},\text{pos}}$, the rigid model best explains the observations as the noise level of this magnitudes hides the underlying model.

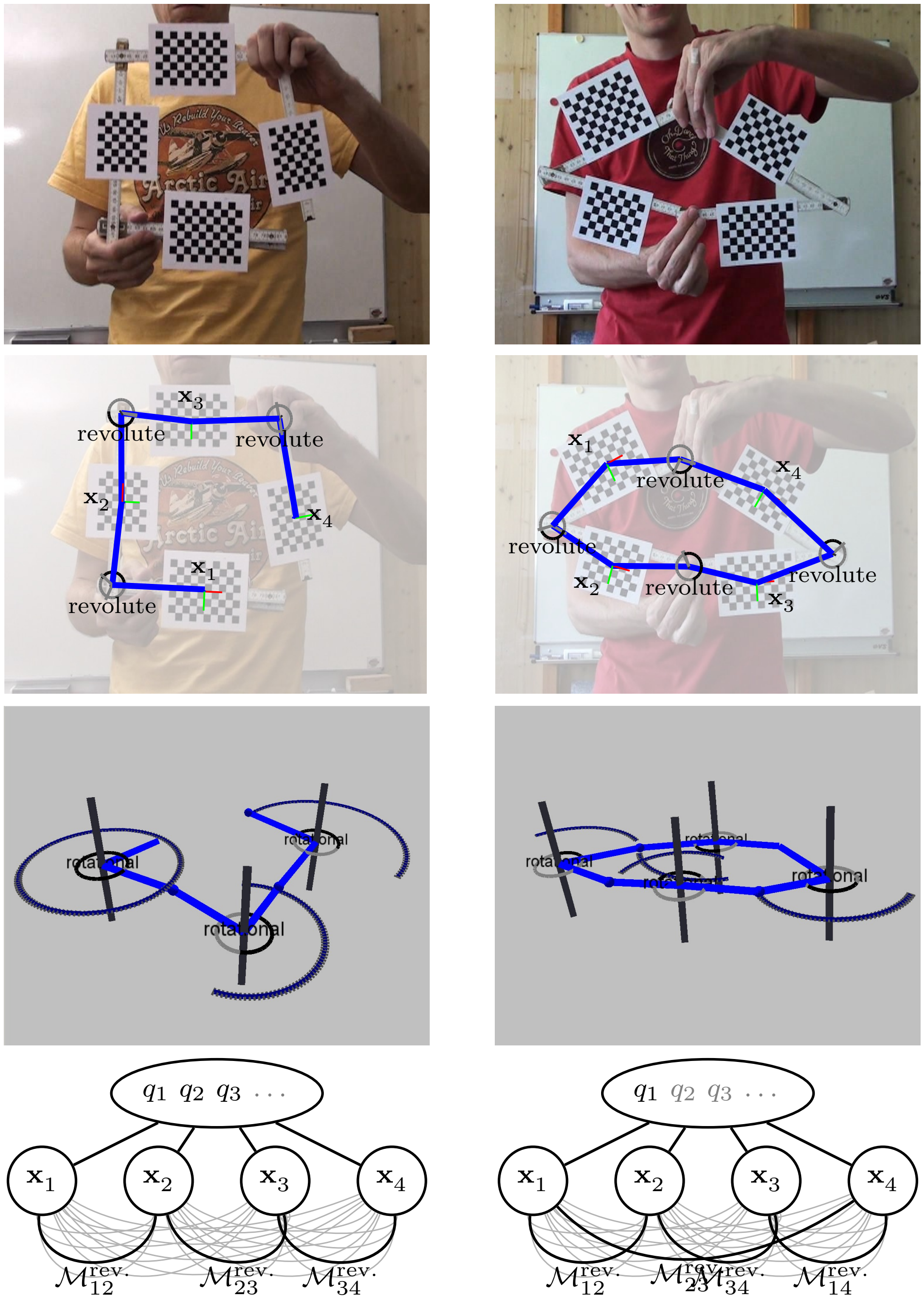

Figure 19: Open kinematic chain with three DOFs (left column) and closed kinematic chain with only a single DOF (right column). First row: images of the objects. Second and third row: learned kinematic models from two different perspectives. Fourth row: learned graphical model, showing the connectivity and the DOFs of the learned kinematic model. The selected kinematic model is visualized by bold edges, the DOFs are given by the boldly type-set configuration variables.

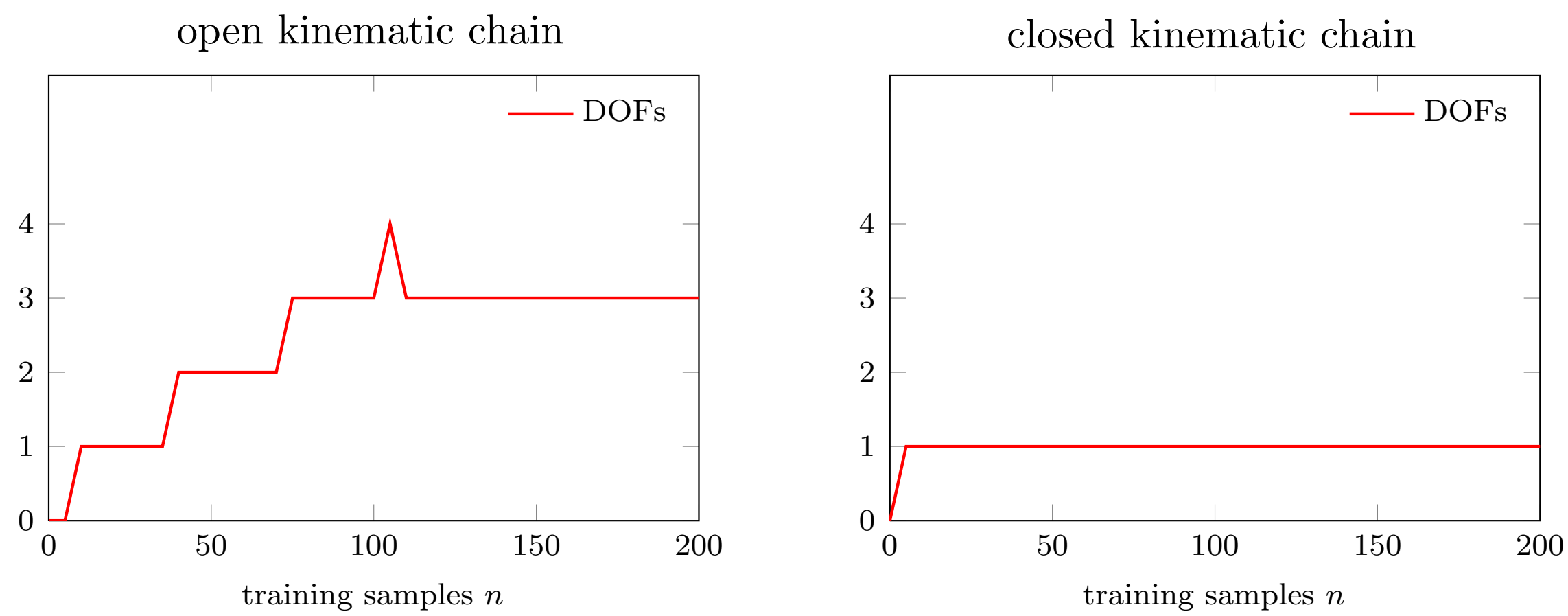

Figure 20: Estimated number of DOFs for the open and the closed kinematic chain object (see Fig. 19). Left: open kinematic chain. Right: closed kinematic chain.

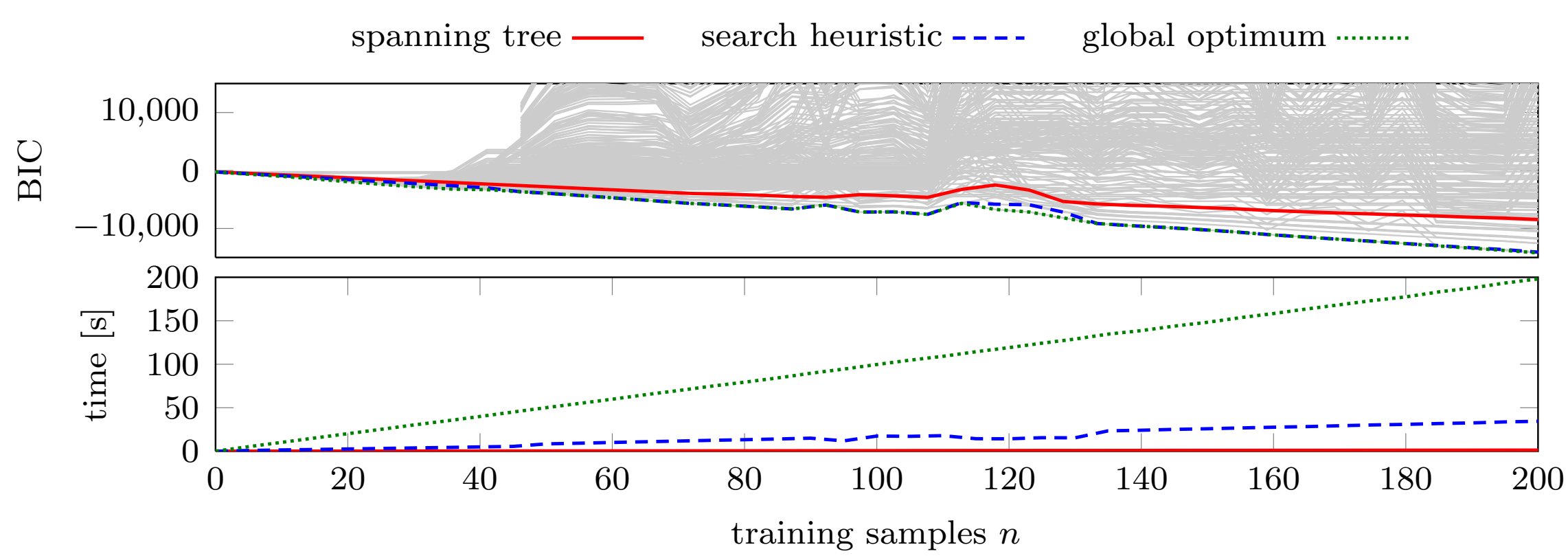

Figure 21: Top: BIC scores of all possible kinematic structures for the closed kinematic chain, as depicted in the top right image of Fig. 20. Bottom: Computation times as a function of the number of training samples.

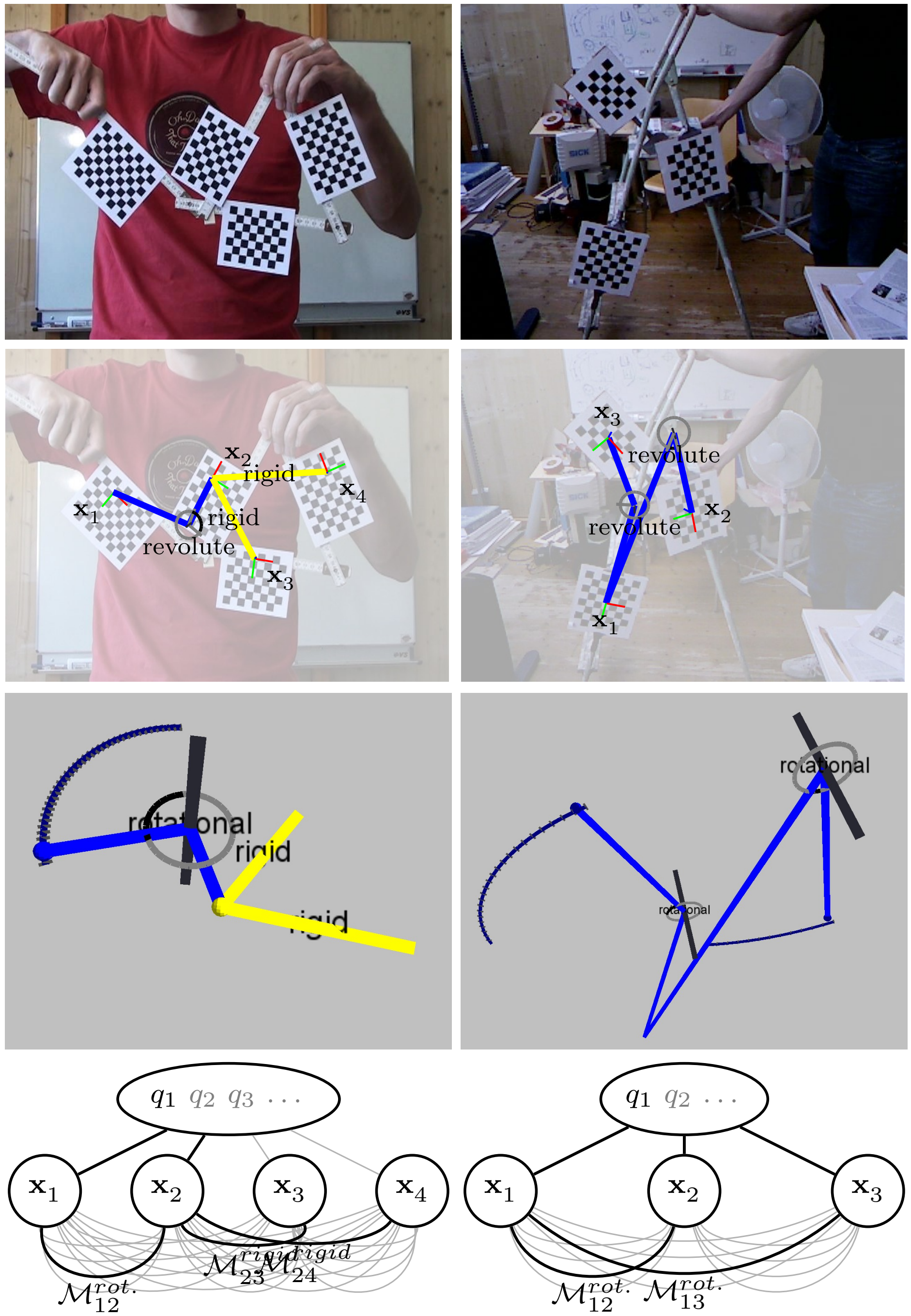

Figure 22: An articulated object consisting of a single revolute joint (left), and a stepladder consisting of two revolute joints (right).

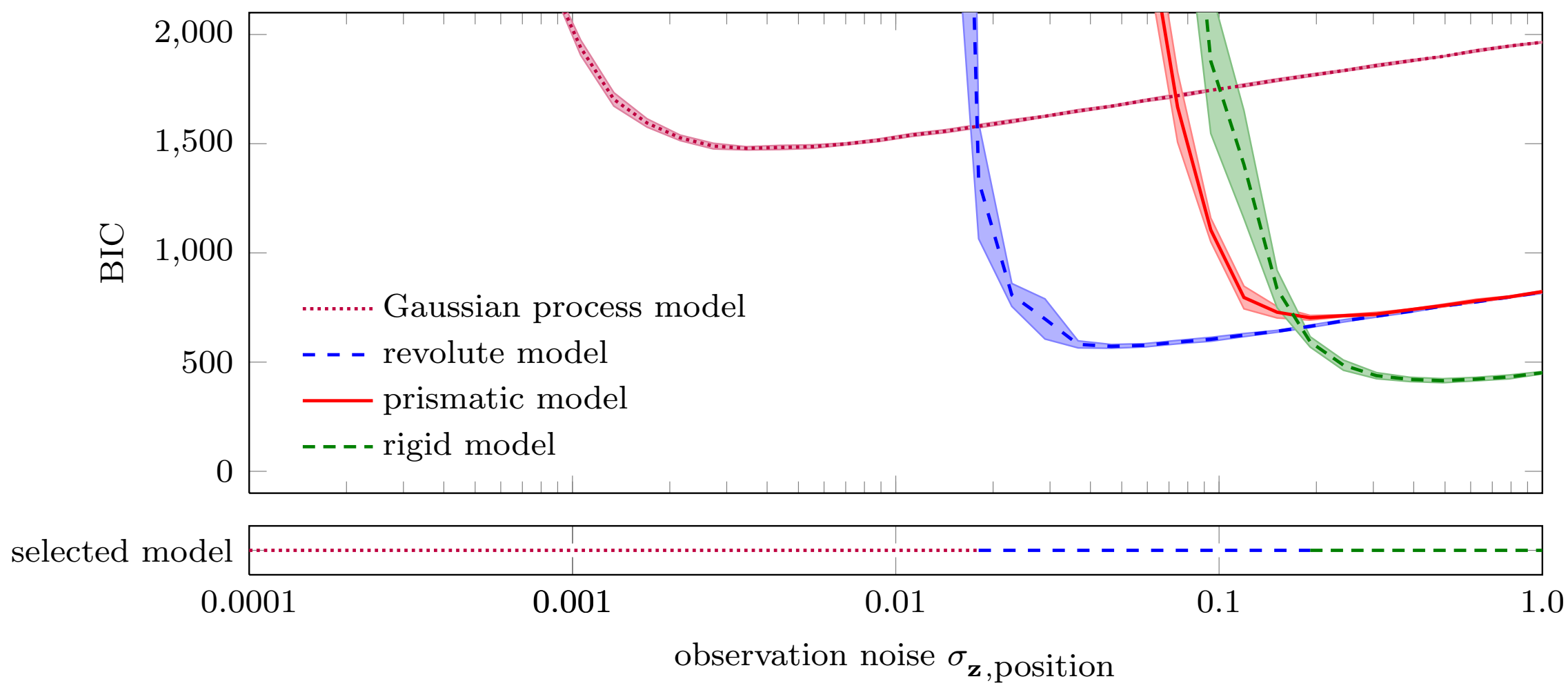

Figure 23: BIC score as a function of the assumed observation noise. A low noise assumption favors the selection of more complex models, and vice versa.

With this experiment, we demonstrated that our model selection procedure is robust over large intervals of the observation noise assumption, i.e., even though the true observation noise was set to $\sigma_{\mathbf{z},\text{pos}}^{\text{true}} = 0.05$, our approach selected the revolute model when the noise assumption was in between $0.02 \leq \sigma_{\mathbf{z},\text{pos}} \leq 0.2$, thus robust over a whole magnitude. We also performed more experiments on synthetic data to verify that our estimators are robust against normally distributed noise and that the MLESAC-based estimators are additionally robust against uniformly distributed outliers.

### 5.6 Open-Source Availability

The source code, documentation, code samples, and tutorials are fully available as open-source, licensed under BSD. We also provide a step-by-step guide to repeat these experiments using a consumer-grade laptop and a webcam[4].

## 6. Related Work

In this paper, we have combined several techniques that come from different fields, i.e., system identification for fitting kinematic models, information theory for model comparison and structure selection, computer vision for estimating and tracking objects, service robotics and control for actually manipulating articulated objects with mobile manipulators. In the

4. http://www.ros.org/wiki/articulation

following, we will review related approaches, contrast them with our approach, and highlight our contributions.

### 6.1 Kinematic Model Fitting

Calibrating kinematic models of manipulation robots to sensor data has a long history in system identification. A good overview of existing techniques can be found in the work of Hollerbach, Khalil, and Gautier (2008). He, Zhao, Yang, and Yang (2010) recently analyzed the identifiability of parameters for serial-chain manipulators, and proposed a generic approach for calibration. Pradeep, Konolige, and Berger (2010) recently presented a system implementation of sensor-actuator calibration for a complex service robot consisting of two arms, a laser scanner and several cameras. In all of these works, the kinematic model is specified in advance, and it is typically expected that an initial parameter set is available. By taking multiple pose observations of the robot in different configurations, the error between prediction and observation can be computed, and finally the parameter vector can optimized by using non-linear, iterative least-squares methods.

In our case, neither the kinematic model nor an initial parameter set is available, but needs to be estimated from the observations alone. In particular when the observations are disturbed by noise and outliers, sample consensus methods have been proven to provide robust estimates (Fischler & Bolles, 1981; Torr & Zisserman, 2000; Nistér, 2005; Rusu, Marton, Blodow, Dolha, & Beetz, 2008). In our model estimators, we use MLESAC as first described by Torr and Zisserman (2000) which is – in contrast to least-squares fitting – robust against reasonable amounts of outliers in the training data.

### 6.2 Kinematic Structure Selection

Estimating kinematic structure from observations has been studied intensively before, however, without subsequently using these models for robotic manipulation (Taycher, Fisher, & Darrell, 2002; Kirk, O'Brien, & Forsyth, 2004; Yan & Pollefeys, 2006; Ross, Tarlow, & Zemel, 2008; Pekelny & Gotsman, 2008). Taycher et al. (2002) address the task of estimating the underlying topology of an observed articulated body. Their focus lies on recovering the topology of the object rather than on learning a generative model. Also, compared to their work, our approach can handle links with more complex link models, e.g., multiple DOFs and non-parametric models. Kirk et al. (2004) extract human skeletal topologies using 3D markers from a motion capture system, however assuming that all joints are revolute. Yan and Pollefeys (2006) present an approach for learning the structure of an articulated object from feature trajectories under affine projections. Other researchers have addressed the problem of identifying different object parts from image data. Ross et al. (2008) use multi-body structure from motion to extract links from an image sequence and then fit an articulated model to these links using maximum likelihood learning.

There exist several approaches where tracking articulated objects is the key motivation and often an a-priori model is assumed. Krainin, Henry, Ren, and Fox (2010), for example, described recently an approach for tracking articulated objects such as a manipulator using a depth camera with a texture projector. However, they require a geometric model of the manipulator. Kragic, Petersson, and Christensen (2002) describe an integrated navigation system for mobile robots which includes a vision-based system for the detection of door handles that enables the robot to successfully open doors. Anguelov, Koller, Parker, and Thrun (2004) model doors as line segments that rotate around a hinge. EM is then used to find the model parameters both from 2D range data and images. Nieuwenhuisen, Stückler, and Behnke (2010) describe an approach where a mobile robot increases its localization accuracy by learning the positions of doors.

Although learning the structure of general Bayesian networks has been proven to be NP-complete (Chickering, 1996), many approximate methods have been proposed that can solve the structure search problem efficiently. Such methods include greedy search, iterated hill climbing, genetic algorithms and ant colony optimization (Chickering, 2002; Daly & Shen, 2009). In some cases, the size of the search space can be reduced significantly by evaluating a number of statistical independence tests (Margaritis & Thrun, 1999; Bromberg, Margaritis, & Honavar, 2009). In this paper, we consider special Bayesian networks representing kinematic structures. This allows us to exploit the mutual independence of the edges of kinematic trees for efficiently recovering the kinematic structure. For closed kinematic chains, we use a greedy search heuristic similar as in the work of Chickering (2002).

Estimating both the structure and the models requires to trade off data fit with model complexity. This corresponds to a Bayesian model selection problem, as described by MacKay (2003). We approach this problem in our work using the Bayesian Information Criterion (BIC) introduced by Schwarz (1978). The BIC provides a method for selecting between alternate model hypotheses, based on their data likelihood and model complexity. In this paper, we use the BIC both to select the kinematic models for the individual links as well as for multi-part articulated objects.

### 6.3 Pose Estimation

Many approaches for estimating the pose of objects from sensory data have been proposed in the past, but solving the general problem is still an ongoing research effort. Marker-based approaches using active or passive markers have the advantage of being easy to use and providing full 3D pose information, but require artificial markers to be attached upon the object parts of interest. Early work in the area of articulated object tracking was presented by Lowe (1991) under the assumption that the object model (and a good initialization) is known. Nieuwenhuisen et al. (2010) use a 2D laser range finder for detecting doors in an office environment and storing them on a map. Tilting lasers or line stripe systems provide dense 3D point clouds and have been used for localizing doors and door handles, but cannot deal with moving objects (Rusu, Meeussen, Chitta, & Beetz, 2009; Quigley, Batra, Gould, Klingbeil, Le, Wellman, & Ng, 2009). Camera-based approaches can provide higher frame rates. A good survey on the state-of-the-art in camera-based pose estimation techniques can be found in the work of Lepetit and Fua (2005). In our context, the work of Murillo, Kosecka, Guerrero, and Sagues (2008) and Andreopoulos and Tsotsos (2008) on visual door detection and pose estimation is of particular relevance. Stereo systems that employ matching algorithms to produce dense results provide 3D point clouds at video frame rates, but suffer from occasional dropouts in areas with low texture or illumination (Konolige, 1997; Brox, Rosenhahn, Gall, & Cremers, 2010; Wedel, Rabe, Vaudrey, Brox, Franke, & Cremers, 2008). This can be overcome by active camera systems that add texture to the scene using a projector LED. Two examples of such systems have been described by Konolige (2010) and Fox and Ren (2010).

In our work, we use several different approaches for estimating the pose of an articulated object for showing that our approach is not specific to a specific data source. In particular, we use marker-based pose estimation from monocular camera, marker-less pose estimation from stereo data, and proprioceptive tracking using the robot's joint encoders.

### 6.4 Operating Articulated Objects

Several researchers have addressed the problem of operating articulated objects with robotic manipulators. A large number of these techniques have focused on handling doors and drawers (Klingbeil et al., 2009; Kragic et al., 2002; Meeussen et al., 2010; Petrovskaya & Ng, 2007; Parlitz, Hägele, Kleint, Seifertt, & Dautenhahn, 2008; Niemeyer & Slotine, 1997; Andreopoulos & Tsotsos, 2008; Rusu et al., 2009; Chitta, Cohen, & Likhachev, 2010). The majority of these approaches, however, assumes an implicit kinematic model of the articulated object. Meeussen et al. (2010) describe an integrated navigation system for mobile robots including vision- and laser-based detection of doors and door handles that enables the robot to successfully open doors using a compliant arm. Diankov, Srinivasa, Ferguson, and Kuffner (2008) formulate door and drawer operation as a kinematically constrained planning problem and propose to use caging grasps to enlarge the configuration space, and demonstrate this on an integrated system performing various fetch-and-carry tasks (Srinivasa, Ferguson, Helfrich, Berenson, Romea, Diankov, Gallagher, Hollinger, Kuffner, & Vandeweghe, 2010). Wieland et al. (2009) combine force and visual feedback to reduce the interaction forces when opening kitchen cabinets and drawers. In contrast to our work, these approaches make strong assumptions on the articulated objects, and do not deal with the problem of inferring their kinematic structure. Therefore, they neither deal with unknown objects, nor improve their performance through learning. Katz and Brock (2008) have enabled a robot to first interact with a planar kinematic object on a table in order to visually learn a kinematic model, and then manipulate the object using this model to achieve a goal state. In contrast to our work, their approach assumes planar objects and learns only 2D models. Jain and Kemp (2009b, 2010) recently presented an approach that enabled a robot to estimate the radius and location of the axis for rotary joints that move in a plane parallel to the ground, while opening novel doors and drawers using equilibrium point control. Recently, we combined (in collaboration with Jain and Kemp) the model learning approach with the equilibrium point controller (Sturm et al., 2010). This enabled the robot to operate a larger class of articulated objects, i.e., objects with non-vertical rotation axes.

## 7. Conclusion

In this paper, we presented a novel approach for learning kinematic models of articulated objects. Our approach infers the connectivity of rigid parts that constitute the object including the articulation models of the individual links. To model the links, our approach considers both, parametrized as well as parameter-free representations. In extensive studies on synthetic and real data, we have evaluated the behavior of model estimation, model selection, and structure discovery. We have shown that our approach is applicable to a wide range of articulated objects, and that it can be used in conjunction with a variety of different sensor modalities. Our approach enables mobile manipulators to operate unknown articulated objects, learn their models, and improve over time.

Despite the promising results presented in this paper, there are several open research questions that remain for future investigation. In our current approach, we learn the kinematic models from static pose observations. It would be interesting to include the velocities or accelerations of object or body parts. This would allow the robot to learn the dynamic parameters as well and enable it to plan time-optimal motion trajectories. A dynamical model would enable the robot to accurately execute motions at higher speeds. Furthermore, a robot that can measure forces and torques while actuating an object could additionally learn friction and damping profiles and include this information in the learned model as well. The robot could benefit from this information to assess, for example, whether a door or drawer is jammed.

## 8. Acknowledgments

The authors gratefully acknowledge the help of Advait Jain and Charlie Kemp from Georgia Tech, in particular for collaboration and joint development of the online model estimation and control approach as described in Section 4.3, and for evaluating the approach on their mobile manipulation robot Cody as described in Section 5.3. Further, the authors would like to thank Vijay Pradeep and Kurt Konolige from Willow Garage who inspired the authors to work on this subject, and contributed to the experiments with the motion capture device as reported in Section 5.1. Additional thanks go to Kurt Konolige for the joint development of the marker-less perception algorithm from stereo data as outlined in Section 4.2 as well as the evaluation presented in Section 5.2. This work has partly been supported by the European Commission under grant agreement numbers FP7-248258-First-MM, FP7-260026-TAPAS, FP7-ICT-248873-RADHAR, and by the DFG under contract number SFB/TR-8.